\documentclass[letterpaper,twocolumn,10pt]{article}
\usepackage{usenix}
\usepackage[available,functional,reproduced]{usenixbadges}
\usepackage{fancyhdr}
\usepackage{flushend}
\usepackage{xspace}
\usepackage{tikz}
\usepackage{subcaption}
\usepackage{booktabs} % for easy table generation
\usepackage{graphics} % for adjusting table width
\usepackage{filecontents}
\usepackage{multirow}
\usepackage{pdfprivacy}
\usepackage{xcolor}
\usepackage{amsmath}
\usepackage{array}
\usepackage{amsfonts}
\usepackage{algorithmic}
\usepackage{textcomp}

\usepackage[font=small,skip=5pt]{caption}
\newcommand{\name}{{InfiniGen\xspace}}
\newcommand{\ho}{{H$_2$O}}

\newcommand{\putsec}[2]{\section{#2}\label{sec:#1}}
\newcommand{\putssec}[2]{\subsection{#2}\label{ssec:#1}}

\newcommand{\secref}[1]{Section~\ref{sec:#1}}
\newcommand{\ssecref}[1]{Section~\ref{ssec:#1}}

\newcommand{\figref}[1]{Figure~\ref{fig:#1}}
\newcommand{\tabref}[1]{Table~\ref{tab:#1}}
\newcommand{\eqnref}[1]{Equation~\ref{eqn:#1}}

\newcommand{\mm}[2]{$\mathrm{\texttt{#1}_{#2}}$}

\newcommand{\myparagraph}[1]{\vspace{0.02in}\noindent\textbf{#1}}

\begin{document}

\date{}

\title{\Large \bf \name{}: Efficient Generative Inference of Large Language Models with \\ Dynamic KV Cache Management}

\author{
{\rm Wonbeom Lee$^{\dag}$ \quad\quad Jungi Lee$^{\dag}$ \quad\quad Junghwan Seo \quad\quad Jaewoong Sim \vspace{0.3em}} \\ 
{Seoul National University}
}

\maketitle
\def\thefootnote{\dag}\footnotetext{Equal contribution}\def\thefootnote{\arabic{footnote}}

\begin{abstract}
  Transformer-based large language models (LLMs) demonstrate impressive
  performance across various natural language processing tasks.  Serving LLM
  inference for generating long contents, however, poses a challenge due to the
  enormous memory footprint of the transient state, known as the key-value (KV)
  cache, which scales with the sequence length and batch size.  
  In this paper, we present \name{}, a novel KV cache management framework
  tailored for long-text generation, which synergistically works with modern
  offloading-based inference systems.  
  \name{} leverages the key insight that a few important tokens that are
  essential for computing the subsequent attention layer in the Transformer can
  be speculated by performing a minimal rehearsal with the inputs of the
  current layer and part of the query weight and key cache of the subsequent
  layer.
  This allows us to prefetch only the essential KV cache entries (without
  fetching them all), thereby mitigating the fetch overhead from the host
  memory in offloading-based LLM serving systems.
  Our evaluation on several representative LLMs shows that \name{} improves the
  overall performance of a modern offloading-based system by up to 3.00$\times$
  compared to prior KV cache management methods while offering substantially
  better model accuracy.
\end{abstract}

\putsec{intro}{Introduction}
Large language models (LLMs) have opened a new era across a wide range of
real-world applications such as chatbots~\cite{zha:emi18,maz:sam18}, coding
assistants~\cite{nij:bo22,che:jer21}, language
translations~\cite{hao:you19,deepl}, and document
summarization~\cite{zha:jia19,wan:zha23}. The remarkable success of LLMs can
largely be attributed to the enormous model size, which enables effective
processing and generation of long contents.
For instance, while the maximum sequence length of the first version of GPT was
restricted to 512 tokens~\cite{rad:nar18}, the latest version, GPT-4, can
handle up to 32K tokens, which is equivalent to approximately 50 pages of
text~\cite{gpt4}.
Some recently announced models such as Claude 3~\cite{claude} and Gemini
1.5~\cite{gemini} can even process up to 1 million tokens, significantly
expanding the context window by several orders of magnitude.

In addition to the well-studied challenge of the model size, deploying LLMs now
encounters a new challenge due to the substantial footprint of the transient
state, referred to as the \emph{key-value (KV) cache}, during long context
processing and generation.
For generative LLM inference, the keys and values of all preceding tokens are
\emph{stored} in memory to avoid redundant and repeated computation. Unlike the
model weights, however, the KV cache scales with the output sequence length,
often consuming even more memory capacity than the model weights. As the demand
for longer sequence lengths (along with larger batch sizes) continues to grow,
the issue of the KV cache size will become more pronounced in the future.

Meanwhile, modern LLM serving systems support offloading data to the CPU memory
to efficiently serve LLMs within the hardware
budget~\cite{ami:raj22,she:zhe23}.
These offloading-based inference systems begin to support even offloading the
KV cache to the CPU memory, thereby allowing users to generate much longer
contexts beyond the GPU memory capacity. However, transferring the massive size
of the KV cache from the CPU memory to the GPU becomes a new performance
bottleneck in LLM inference.

In this work, we propose \name{}, a KV cache management framework designed to
synergistically work with modern offloading-based inference systems. 
\name{} builds on two key design principles.
First, it speculates and chooses the KV cache entries that are critical to
produce the next output token, dropping the non-critical ones, by conducting a
minimal \emph{rehearsal} of attention computation for Layer $i$ at Layer $i-1$.
Second, it leverages the CPU memory capacity and maintains the KV cache pool on
the CPU, rather than on the GPU, to ensure that the critical KV cache values
can be identified for all outputs and layers with a large \emph{window size}
while alleviating the concerns about limited GPU memory capacity for long
content generation.

In particular, \name{} manipulates the model weights \emph{offline} to make the
speculation far more efficient and precise, by skewing the Transformer
architecture query and key matrices to emphasize certain important columns.
During the prefill stage, while the prompt and input of an inference request
are initially processed, \name{} generates \emph{partial} weights for use in
the subsequent decoding (i.e., output generation) stage. 
At Layer $i-1$ of the decoding stage, \name{} speculates on the attention
pattern of the next layer (Layer $i$) using the attention input of Layer $i-1$,
a partial query weight, and a partial key cache of Layer $i$.
Based on the speculated attention pattern, \name{} prefetches the essential KV
cache entries from the CPU memory for attention computation at Layer $i$.
By dynamically adjusting the number of KV entries to prefetch, \name{} brings
only the necessary amount of the KV cache to the GPU, thereby greatly reducing
the overhead of the KV cache transfer. In addition, \name{} manages the KV
cache pool by dynamically removing the KV cache entries of infrequently used
tokens. 

We implement \name{} on a modern offloading-based inference
system~\cite{she:zhe23} and evaluate it on two representative LLMs with varying
model sizes, batch sizes, and sequence lengths.
Our evaluation shows that \name{} achieves up to a 3.00$\times$ speedup over
the existing KV cache management methods while offering up to a 32.6 percentage
point increase in accuracy.
In addition, \name{} consistently provides performance improvements with larger
models, longer sequence lengths, and larger batch sizes, while prior
compression-based methods lead to saturating speedups.

In summary, this paper makes the following contributions: 
\begin{itemize}
  \item We present \name{}, a dynamic KV cache management framework that
    synergistically works with modern offloading-based LLM serving systems by
    intelligently managing the KV cache in the CPU memory.
  \item We propose a novel KV cache prefetching technique with {ephemeral}
    pruning, which speculates on the attention pattern of the subsequent
    attention layer and brings only the essential portion of the KV cache to
    the GPU while retaining the rest in the CPU memory.
  \item We implement \name{} on a modern offloading-based inference system and
    demonstrate that it greatly outperforms the existing KV cache management
    methods, achieving up to 3.00$\times$ faster performance while also
    providing better model accuracy.
\end{itemize}

\putsec{back}{Background}
This section briefly explains the operational flow and the KV caching technique
of large language models and introduces the singular value decomposition (SVD)
as a method of skewing matrices for a better understanding of our proposed
framework, which we discuss in~\secref{base-algo}.

%%%%%%%%%%%%%%%%%%%%%%%%%%%%%%%%%%%%%%%%%%%%%%%%%%%%%%%%%%%%%%%%%%%%%%%%
\putssec{llm}{Large Language Models}
Large language models (LLMs) are composed of a stack of Transformer blocks,
each of which contains an attention layer followed by a feed-forward
layer~\cite{vas:sha17}.
The input tensor ($X$) of the Transformer block has a dimension of $N \times
D$, where $N$ is the number of query tokens, and $D$ is the model dimension.
This input tensor ($X$) is first layer-normalized (LayerNorm), and the
layer-normalized tensor ($X_a$) is fed into the attention layer as input.
The attention input ($X_a$) is multiplied by three different \emph{weight}
matrices ($W_Q$, $W_K$, $W_V$) to generate Query ($Q$), Key ($K$), and Value
($V$) matrices.  
Each weight matrix has a dimension of $D \times D$. Thus, Query, Key, and Value
have a dimension of $N \times D$. These matrices are reshaped to have a
dimension of $H \times N \times d$, where $H$ is the number of attention heads
and $d$ is the head dimension; note that $D = H \times d$.  

Each head individually performs attention computation, which can be formulated
as follows: $\mathrm{softmax}(QK^T)V$.\footnote{In this work, we refer to the
results of $QK^T$ and $\mathrm{softmax}(QK^T)$ as \emph{attention scores} and
\emph{attention weights}, respectively.}
The attention output, after a residual add (adding to the input tensor $X$) and
layer normalization, is fed into the feed-forward layer. The feed-forward
network (FFN) consists of two consecutive linear projections and a non-linear
activation operation between them. The output of FFN after a residual add
becomes the output of a Transformer block, which has the \emph{same}
dimensionality as the input of the Transformer block (i.e., $N \times D$). This
allows us to easily scale LLMs by adjusting the number of Transformer blocks.

%%%%%%%%%%%%%%%%%%%%%%%%%%%%%%%%%%%%%%%%%%%%%%%%%%%%%%%%%%%%%%%%%%%%%%%%
\putssec{inference}{Generative Inference and KV Caching}
Generative LLM inference normally involves two key stages: the \emph{prefill}
stage and the \emph{decoding} stage.
In the prefill stage, LLMs summarize the context of the input sequence (i.e.,
input prompt) and produce a \emph{new} token that serves as the initial input
for the decoding stage.
Subsequently, using this new token, LLMs run the decoding stage to generate the
next token. The newly generated token is then fed back into the decoding stage
as input, creating an \emph{autoregressive} process for token generation. In
this work, we refer to each token generation in the decoding stage as an
\emph{iteration}.

To generate a new token that aligns well with the context, LLMs need to compute
the relationship between the last token and all the previous ones, including
the tokens from the input sequence, in the attention layer.
A na\"ive approach to this is to recompute the keys and values of all the
previous tokens at every iteration.  However, this incurs a significant
overhead due to redundant and repeated computation.
Furthermore, the computation overhead linearly grows with the number of the
previous tokens; i.e., the overhead becomes larger for longer sequences.

To avoid such overhead, the keys ($K$) and values ($V$) of all the previous
tokens are typically \emph{memoized} in memory, which is known as the \emph{KV
cache}. The KV cache then keeps updated with the key and value of the generated
token at each iteration. 
As such, the dimension of the KV cache at the $i$-th iteration can be expressed
as $H \times (N + i) \times d$.
If batched inference is employed, the size of the KV cache also grows linearly
to the batch size.
By employing the KV cache, we can avoid repeated computation and produce the
key and value of only one token at each iteration.
Note that in the decoding stage, the input to the Transformer block ($X$) has a
dimension of $1 \times D$, and the dimension of the attention score matrix
becomes $H \times 1 \times (N + i)$ at the $i$-th iteration.

%%%%%%%%%%%%%%%%%%%%%%%%%%%%%%%%%%%%%%%%%%%%%%%%%%%%%%%%%%%%%%%%%%%%%%%%
\putssec{outlier}{Outliers in Large Language Models}
Large language models have outliers in the Transformer block input tensors. The
outliers refer to the elements with substantially larger magnitudes than the
other elements. The outliers in LLMs appear in a few fixed channels (i.e.,
columns in a 2D matrix) across the layers. Prior work has shown that outliers
are due to the intrinsic property of the model (e.g., large magnitudes in a few
fixed channels of layer normalization weights)~\cite{wei:yun22, det:lew22}.

\begin{figure}[t]
  \includegraphics[width=\linewidth]{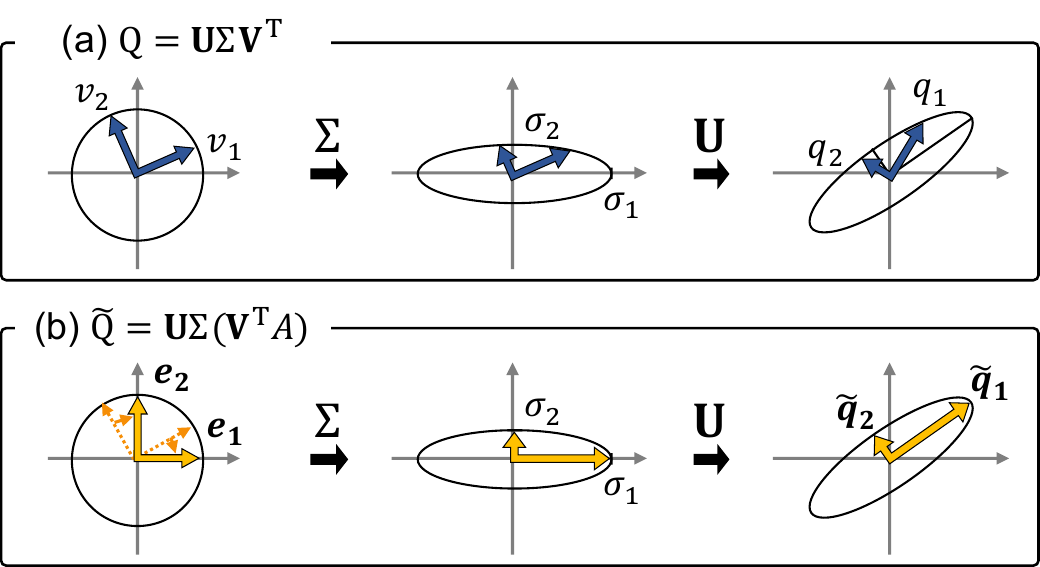}
  \caption{
    Transformation from matrix $\mathbf{V}^{T}$ to matrix $Q$ in terms of SVD.
    The orthogonal matrix $A$ maximizes the difference in magnitude between the
    column vectors of $Q$.
  }
  \vspace{-0.15in}
  \label{fig:svd}
\end{figure}

%%%%%%%%%%%%%%%%%%%%%%%%%%%%%%%%%%%%%%%%%%%%%%%%%%%%%%%%%%%%%%%%%%%%%%%%
\putssec{svd}{Singular Value Decomposition}
We observe that skewing the query and key matrices to make a small number of
channels much larger than others and using only those channels to compute the
attention score matrix can effectively predict which tokens are important. In
essence, we multiply the $Q$ and $K$ matrices with an orthogonal matrix $A$ to
make it align with the direction that $Q$ stretches the most, to produce the
respective skewed matrices $\Tilde{Q}$ and $\tilde{K}$. We explain in detail
why we use an orthogonal matrix in~\ssecref{opportunity}.

To find such an orthogonal matrix $A$, we employ the singular value
decomposition (SVD), which is a widely used matrix factorization technique in
linear algebra. 
For a real matrix $Q$ of size $m \times n$, its SVD factorization can be
expressed as follows:
\begin{equation}
  \begin{aligned}
    Q = \mathbf{U} \, \Sigma \, \mathbf{V}^{T}, \nonumber
    \label{eqn:svd}
  \end{aligned}
\end{equation}
where $\mathbf{U}$ and $\mathbf{V}$ are orthogonal matrices of size $m \times
m$ and $n \times n$, respectively.\footnote{Note that this $\mathbf{V}$,
typeset with a different font, is one of the resulting matrices of SVD and is
distinct from the $V$ of the Value matrix in the Transformer attention layer.}
$\Sigma$ is an $m \times n$ diagonal matrix, which has nonzero values
($\sigma_{1},\sigma_{2}, ..., \sigma_{k}$) on the diagonal, where
$k=\mathrm{min}({m,n})$.
In terms of linear transformation, it is well known that a transformation of a
vector $v \in \mathbb{R}^{n}$ by a real matrix $B$ (i.e., the product of $B$
and $v$) is a rotation and/or reflection in $\mathbb{R}^{n}$ if the $B$ matrix
is orthogonal. 
If $B$ is an $m \times n$ diagonal matrix, each dimension of $v$ is stretched
by the corresponding diagonal entry of $B$ and is projected to
$\mathbb{R}^{m}$. 

For example, \figref{svd} shows how the column vectors $v_{1}$ and $v_{2}$ of
$\mathbf{V}^{T}$ would transform to column vectors $q_{1}$ and $q_{2}$ of $Q$,
when $m$ and $n$ are 2. 
In \figref{svd}(a), the orthogonal unit vectors $v_{1}$ and $v_{2}$ are first
stretched to the points on an ellipse whose semi-axis lengths correspond to the
diagonal entries in $\Sigma$. 
The vectors are then rotated and/or reflected to $q_{1}$ and $q_{2}$ by matrix
$\mathbf{U}$. On the other hand, \figref{svd}(b) shows how orthogonal matrix
$A$ performs rotation to make the resulting $\Tilde{q_{1}}$ much larger than
$\Tilde{q_{2}}$. 
Specifically, $A$ rotates vectors $v_{1}$ and $v_{2}$ to $e_{1}$ and $e_{2}$,
which map to the semi-axes of the ellipse. In this way, the vectors are
stretched to the maximum and minimum by the matrix $\Sigma$.
This process emphasizes the magnitude of $\Tilde{q_{1}}$ over $\Tilde{q_{2}}$,
which allows us to effectively predict the attention score using only
$\Tilde{q_{1}}$ while omitting $\Tilde{q_{2}}$.

\putsec{motiv}{Motivation}
In this section, we first explain that the KV cache size becomes a critical
issue for long-text generation in LLM inference, and it becomes more
problematic when deploying modern offloading-based inference
systems~(\ssecref{kv_size}). 
We then discuss why the existing KV cache management methods cannot
fundamentally address the problem in the offloading-based inference
system~(\ssecref{kv_eviction}).

\begin{figure}[t]
  \includegraphics[width=\linewidth]{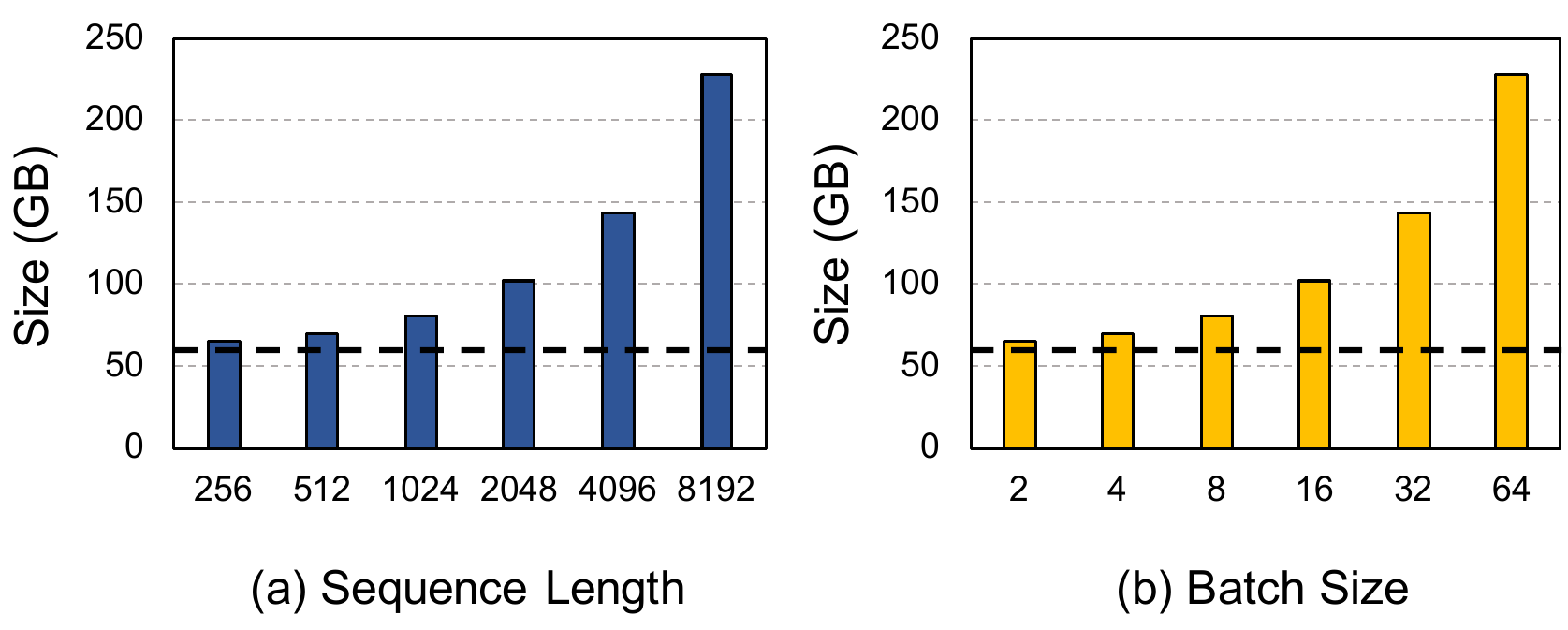}
  \caption{
    Total size of the KV cache and model weights of OPT-30B for different
    sequence lengths and batch sizes. The batch size of (a) is 16, and the
    sequence length of (b) is 2048. The dotted line represents the size of the
    model weights.
  }
  \vspace{-0.15in}
  \label{fig:kv_size}
\end{figure}

%%%%%%%%%%%%%%%%%%%%%%%%%%%%%%%%%%%%%%%%%%%%%%%%%%%%%%%%%%%%%%%%%%%%%%%%
\putssec{kv_size}{KV Cache in LLM Inference Systems}
As discussed in~\ssecref{inference}, today's LLM serving systems exploit KV
caching to avoid redundant computation of key and value projections during the
decoding stage. While this is an effective solution for short sequence
generation with a single client request, the KV cache quickly becomes a key
memory consumer when we generate long sequences or employ modern request
batching techniques~\cite{yu:jeo22,she:zhe23}.

\figref{kv_size} shows the combined size of LLM weights and the KV cache across
different sequence lengths and batch sizes. As depicted in the figure, the
model size remains constant regardless of sequence lengths or batch sizes,
whereas the KV cache size linearly scales with them. 
Note that modern LLM serving systems, such as NVIDIA Triton Inference
Server~\cite{triton} and TensorFlow Serving~\cite{chr:fan17}, already support
\emph{batched} inference for better compute utilization and higher throughput
in serving client requests. 
When individual requests are batched, each request retains its own KV cache,
thereby increasing the overall KV cache size for the inference.
Even for a single client request, beam search~\cite{sut:vin14} and parallel
sampling~\cite{fan:lew18} are widely used to generate better outputs or to
offer clients a selection of candidates~\cite{copilot,che:jer21}.
The techniques also increase the size of the KV cache like batched inference as
multiple sequences are processed together.
Consequently, the KV cache size can easily exceed the model size for many
real-world use cases, as also observed in prior
work~\cite{zha:she23,she:zhe23,liu:des23,pop:dou23}.
This can put substantial pressure on GPU memory capacity, which is relatively
scarce and expensive.

\begin{figure}[t]
  \includegraphics[width=\linewidth]{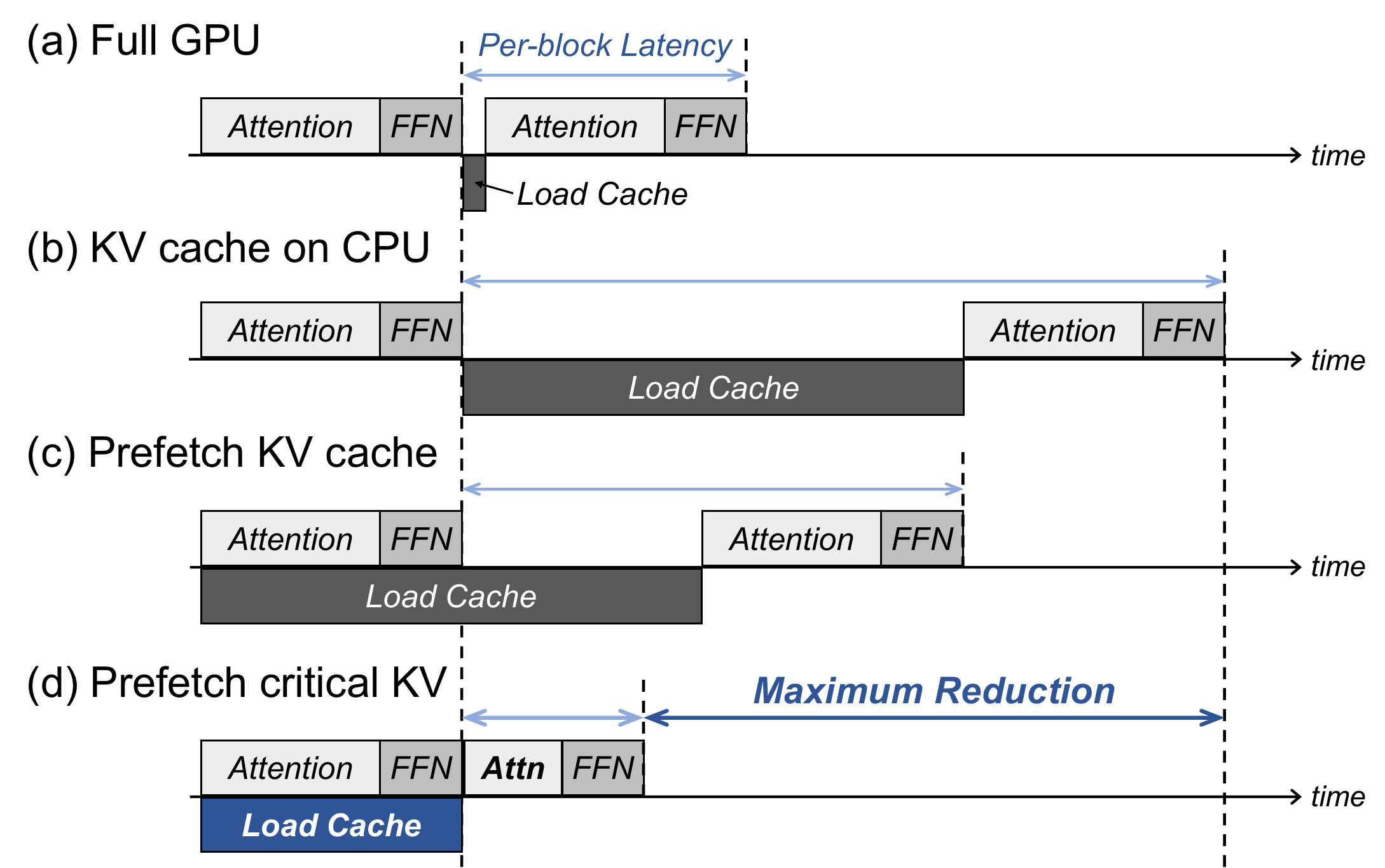}
  \caption{Comparison between different execution styles of Transformer blocks.}
  \vspace{-0.15in}
  \label{fig:block_latency_comparison}
\end{figure}

%%%%%%%%%%%%%%%%%%%%%%%%%%%%%%%%%%%%%%%%%%%%%%%%%%%%%%%%%%%%%%%%%%%%%%%%
\myparagraph{LLM Inference Systems with Offloading.}
Modern LLM serving systems such as DeepSpeed~\cite{ami:raj22} and
FlexGen~\cite{she:zhe23} already support offloading the model weights or the KV
cache to the CPU memory. When it comes to offloading-based inference systems,
the KV cache size becomes more problematic due to the low PCIe bandwidth
between the CPU and GPU, which becomes a new and critical bottleneck.

\figref{block_latency_comparison} depicts a high-level timing diagram between
different execution styles of Transformer blocks.
\figref{block_latency_comparison}(a) represents the case when the KV cache
entirely resides in the GPU memory (Full GPU). In this case, the load latency
of the KV cache (Load Cache) involves a simple read operation from the GPU
memory, which is negligible due to the high bandwidth of GPU memory.  
However, the maximum batch size or sequence length is limited by the GPU memory
capacity, which is relatively smaller than the CPU memory.

To enable a larger batch size or a longer sequence length, we can offload the
KV cache to CPU memory (KV cache on CPU), as shown
in~\figref{block_latency_comparison}(b). 
While offloading-based inference systems alleviate the limitation on the batch
size and sequence length, transferring hundreds of gigabytes of the KV cache to
the GPU for attention computation significantly increases the overall execution
time of Transformer blocks due to the limited PCIe bandwidth.  

Even when we apply a conventional prefetching technique (Prefetch KV cache), as
shown in~\figref{block_latency_comparison}(c), only part of the load latency
can be hidden by the computation of the preceding Transformer block. 
Note that although compressing the KV cache via quantization could potentially
reduce the data transfer overhead in offloading-based systems~\cite{she:zhe23},
it does not serve as a fundamental solution as quantization does not address
the root cause of the KV cache problem, which is the linear scaling of KV
entries with the sequence length.
This necessitates intelligent KV cache management to mitigate the performance
overhead while preserving its benefits.

\begin{figure}[t]
  \includegraphics[width=\linewidth]{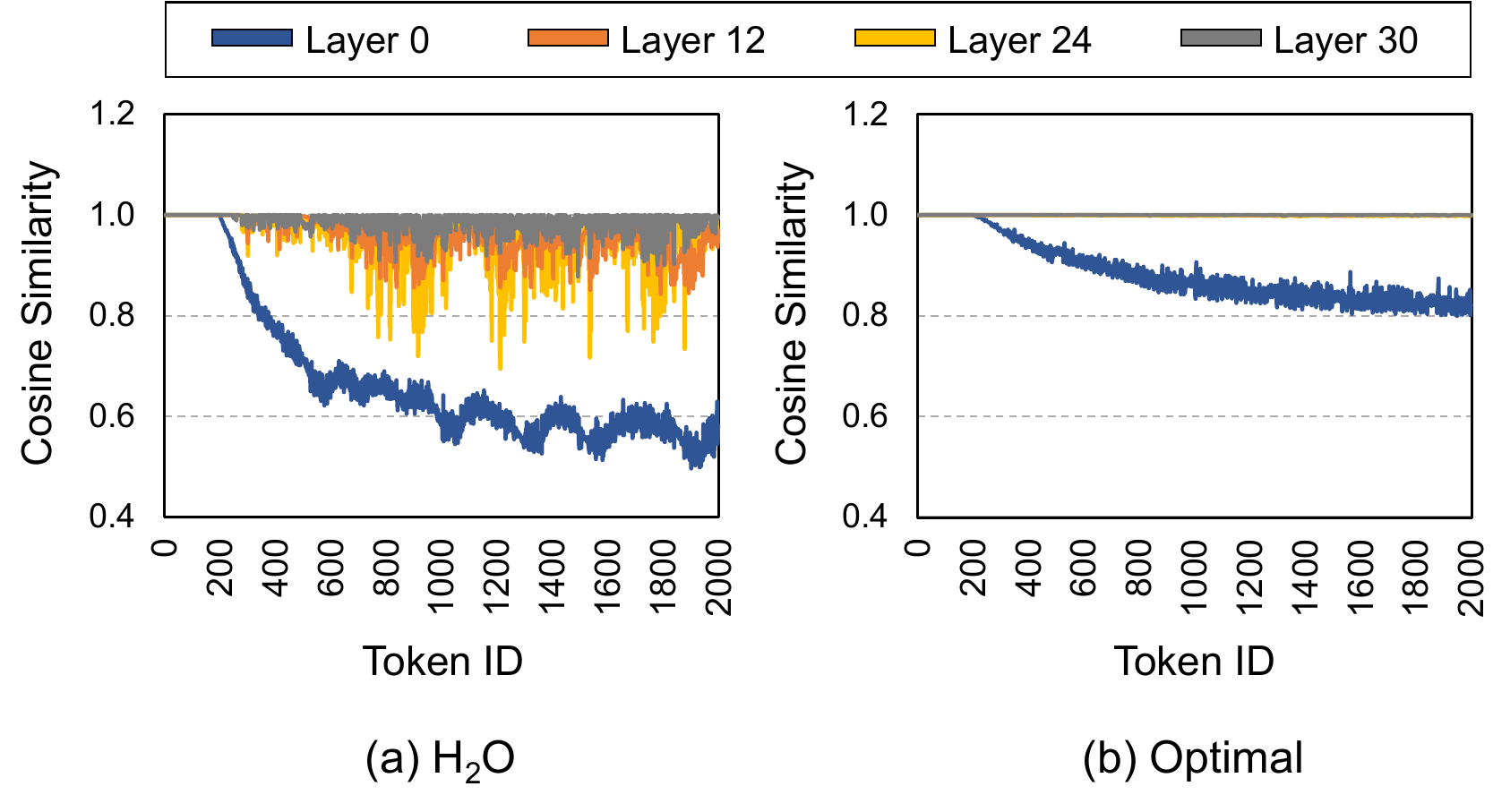}
  \caption{
    Cosine similarity between the attention weights of the base model with full
    cache and (a) \ho{} or (b) Optimal. \ho{} and Optimal use 200 tokens for
    attention computation. We use OPT-6.7B and a random sentence with 2000
    tokens from the PG-19 dataset~\cite{rae:pot20}.  
  }
  \vspace{-0.15in}
  \label{fig:h2o_long_seq}
\end{figure}

%%%%%%%%%%%%%%%%%%%%%%%%%%%%%%%%%%%%%%%%%%%%%%%%%%%%%%%%%%%%%%%%%%%%%%%%
\putssec{kv_eviction}{Challenges in KV Cache Management}
The fundamental approach to mitigating the transfer overhead of the KV cache
from the CPU to GPU is to reduce the volume of the KV cache to load by
identifying the \emph{critical} keys and values for computing attention scores,
as shown in~\figref{block_latency_comparison}(d).
It is widely recognized that the keys and values of certain tokens are more
important than others in attention computation~\cite{kit:kai20, wan:li20,
che:liu21, che:dao21, cho:lik21}. 
As explained in~\ssecref{llm}, after computing the attention score, the softmax
operation is applied, which emphasizes a few large values of tokens. Therefore,
skipping attention computation for some less critical tokens does not
significantly degrade the model accuracy, provided the token selection is
appropriate. 

In this context, several recent works propose to reduce the KV cache size
through key/value evictions at runtime within a constrained KV cache
budget~\cite{liu:des23, zha:she23}. 
However, all the prior works assume the \emph{persistence} of attention
patterns across iterations; that is, if a token is deemed unimportant in the
current iteration (i.e., having a low attention weight), it is likely to remain
unimportant in the generation of future tokens.
Under the assumption, they evict the tokens with a low attention weight from
the KV cache at each iteration when the KV cache size exceeds its budget.
The keys and values of the evicted tokens are \emph{permanently} excluded from
the subsequent iterations while being removed from the memory. 
Although the recent works on managing the KV cache can be applied to
offloading-based inference systems, we observe that they do not effectively
address the challenges in KV cache management below and thus have subpar
performance with offloading-based inference systems.

%%%%%%%%%%%%%%%%%%%%%%%%%%%%%%%%%%%%%%%%%%%%%%%%%%%%%%%%%%%%%%%%%%%%%%%%
\myparagraph{\emph{C1: Dynamic nature of attention patterns across iterations.}}
\figref{h2o_long_seq} shows the cosine similarity between the attention weights
of the baseline model, which uses the KV cache of all prior tokens for
computing attention weights (i.e., a maximum of 2000 tokens in the experiment),
and two different KV cache management methods~(\ho{} and Optimal) with a KV
cache budget of 200 tokens.\footnote{The cosine similarity measures how much
each row of the attention weight is similar to the case of the full KV cache.
If they are similar, the generated tokens will also be similar. Thus, a low
cosine similarity indicates low accuracy far from the baseline model with the
full KV cache.}
\ho{}~\cite{zha:she23} is a state-of-the-art technique that retains only a
small percentage of important tokens in the KV cache to reduce its size. It
\emph{assesses} the importance of each token in every iteration and
\emph{removes} unimportant ones before the next iteration to keep the KV cache
size in check (i.e., using a narrow assessment window). 
In contrast, Optimal represents the scenario where we choose the same number of
tokens as \ho{} from the KV cache at each iteration but retain all prior keys
and values (i.e., using a wider assessment window). In other words, Optimal
selects 200 tokens out of the entire sequence of previous tokens at each
iteration. 

The figure indicates that despite \ho{}-like approaches assuming that the
attention pattern does not change across iterations, this is not the case in
practice. The tokens deemed unimportant in the current iteration could become
important in subsequent iterations. Consequently, \ho{} exhibits high
similarity until around 200 iterations (i.e., within the KV cache budget), but
as the sequence length extends beyond the KV cache budget, it starts to
struggle with the dynamic nature of the attention pattern, resulting in lower
cosine similarity than the Optimal case. 
Note that while we only show the scenario of a KV cache budget of 200 out of a
total sequence length of 2000 tokens for brevity, this issue would become more
pronounced as the sequence length surpasses it.

%%%%%%%%%%%%%%%%%%%%%%%%%%%%%%%%%%%%%%%%%%%%%%%%%%%%%%%%%%%%%%%%%%%%%%%%
\myparagraph{\emph{C2: Adjusting the number of KV entries across layers.}}
\figref{h2o_long_seq} also illustrates that the impact of the KV cache eviction
\emph{varies} across the layers in LLMs.
For Layer 0, both \ho{} and Optimal show a significant drop in cosine
similarity as the token ID increases. This implies that Layer 0 has a
\emph{broader} attending pattern than other layers; i.e., the attention weights
are relatively similar between key tokens. 
Thus, the selected 200 tokens with the large attention weight do \emph{not}
adequately represent the attention pattern of the baseline model for this
layer, as they are likely only slightly larger than the others, not strongly
so. In such cases, it becomes necessary to compute the attention weight with a
larger number of tokens.

\begin{figure}[t]
  \includegraphics[width=\linewidth]{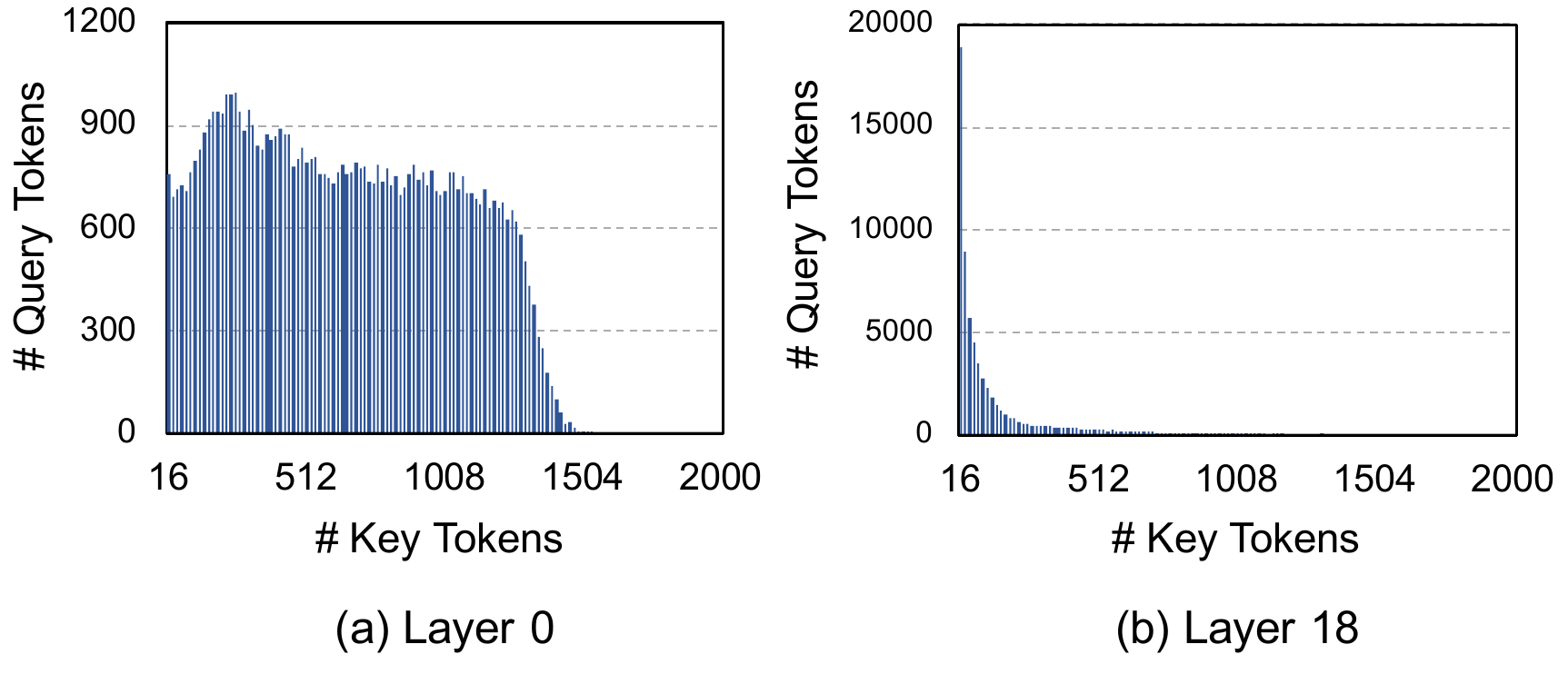}
  \caption{
    Histogram that shows the number of key tokens needed to achieve 0.9 out of
    1.0 total attention weight for (a) Layer 0 and (b) Layer 18 of the OPT-6.7B
    model.  The bin width is set to 16. We observe that the distribution
    dynamically changes across the layers.
  }
  \vspace{-0.15in}
  \label{fig:atten_count_histo}
\end{figure}

To estimate how many keys/values from the KV cache need to be retained, we sort
the attention weight for each query token in descending order and sum the key
tokens until the cumulative weight reaches 0.9.
\figref{atten_count_histo} presents a histogram of the number of query tokens
(y-axis) requiring the number of key tokens (x-axis) needed to reach a weight
of 0.9 (out of the total attention weight of 1.0) in two different layers:
Layer 0 and Layer 18.
Layer 0 shows a broad distribution, indicating a significant variation in the
number of key tokens required to achieve a weight of 0.9 for each query token.
In contrast, Layer 18 exhibits a highly skewed distribution, suggesting that
the majority of the query tokens in this layer require only a few key tokens to
reach a weight of 0.9. 
This implies that we need to \emph{dynamically} adjust the number of key tokens
participating in attention computation across different layers to make
efficient use of the KV cache budget.

%%%%%%%%%%%%%%%%%%%%%%%%%%%%%%%%%%%%%%%%%%%%%%%%%%%%%%%%%%%%%%%%%%%%%%%%
\myparagraph{\emph{C3: Adjusting the number of KV entries across queries.}}
\ho{} sets the number of key/value tokens to retain as a fixed percentage of
the input sequence length. The KV cache budget remains constant regardless of
how many tokens have been generated.
By analyzing the data from~\figref{atten_count_histo} on Layer 18, we observe
that this fixed KV cache budget has some limitations. 
For instance, with an input sequence length of 200 and a 20\% KV cache budget,
\ho{} maintains 40 key/value tokens throughout token generations. 
However, most of the subsequent query tokens require more than 40 tokens to
effectively represent the attention weight of the baseline model; for example,
the 500$^\textrm{th}$, 1000$^\textrm{th}$, 1500$^\textrm{th}$, and
2000$^\textrm{th}$ tokens need 80, 146, 160, and 164 key tokens, respectively,
to reach a cumulative attention weight of 0.9.
This implies an inadequate amount of the key/value tokens to properly represent
the attention weight of the baseline model.
Furthermore, the number of key tokens required to reach 0.9 varies even for the
adjacent query tokens; for instance, the 998$^\textrm{th}$, 999$^\textrm{th}$,
1000$^\textrm{th}$, 1001$^\textrm{st}$, and 1002$^\textrm{nd}$ tokens need 172,
164, 146, 154, and 140 key tokens, respectively. 
Fixing the KV cache budget without accounting for the variance between query
tokens inevitably results in ineffective KV cache management. 
Therefore, we need to \emph{dynamically} adjust the amount of the key/value
tokens loaded and computed for each query token to efficiently manage the KV
cache.

%%%%%%%%%%%%%%%%%%%%%%%%%%%%%%%%%%%%%%%%%%%%%%%%%%%%%%%%%%%%%%%%%%%%%%%%
\myparagraph{Summary.}
Prior works aiming to reduce the KV cache size through token eviction
inherently have some challenges.
Given the dynamic attention pattern across iterations, permanently excluding
evicted tokens from future token generation can result in a non-negligible drop
in accuracy.
Instead, we need to dynamically \emph{select} critical tokens from the KV cache
while avoiding the outright eviction of less important ones.
Furthermore, the fixed size of the KV cache budget in prior works leads to
inefficient KV cache management. The number of key/value tokens required for
each layer differs, and each query token demands a varying number of key/value
tokens to effectively represent the attention pattern of the baseline model. 
Failing to account for these variations may result in ineffective KV cache
management. Thus, we need to dynamically adjust the number of key/value tokens
to select from the KV cache while considering the variances between layers and
query tokens.

\putsec{base-algo}{\name{} Design}
In this section, we present \name{}, a KV cache management framework for
offloading-based inference systems. We first show the high-level overview of
our proposed KV cache management solution~(\ssecref{proposal_overview}) and
discuss the opportunities of KV cache prefetching that we
observe~(\ssecref{opportunity}).
We then explain our prefetching module~(\ssecref{prefetching}), which builds on
the offloading-based inference systems, and discuss how \name{} manages the KV
cache on CPU memory regarding the memory pressure~(\ssecref{management}).

%%%%%%%%%%%%%%%%%%%%%%%%%%%%%%%%%%%%%%%%%%%%%%%%%%%%%%%%%%%%%%%%%%%%%%%%
\putssec{proposal_overview}{Overview}

\begin{figure}[t]
  \centering
  \includegraphics[width=0.95\linewidth]{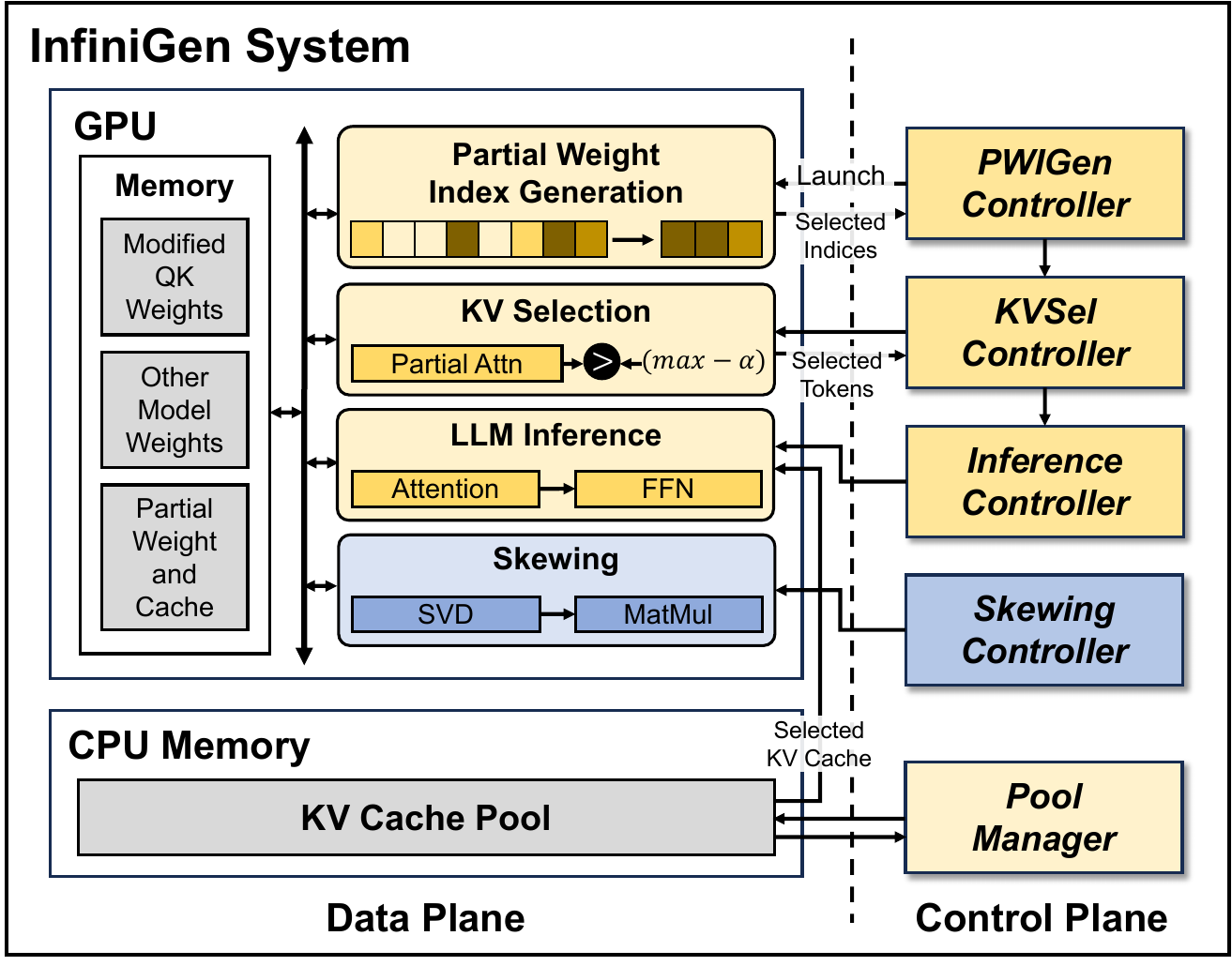}
  \caption{Overview of \name{} design.}
  \vspace{-0.20in}
  \label{fig:framework_diagram}
\end{figure}

\figref{framework_diagram} shows an overview of our KV cache management
framework, \name{}, which enables offloading the KV cache with low data
transfer overhead.
The key design principle behind \name{} is to exploit the abundant CPU memory
capacity to increase the \emph{window size} when identifying the important
tokens in the KV cache.
As such, the majority of the tokens for the KV cache are kept in the CPU memory
as we generate new tokens, not completely discarding them unlike prior
works~\cite{zha:she23,liu:des23}. 
However, we do not bring the entire KV cache to the GPU for attention
computation, but load and compute with only the keys and values of a \emph{few
important tokens}, dropping other unimportant ones dynamically.
To do so, we maintain the KV cache pool in the CPU memory and selectively and
speculatively load a few of tokens.

In detail, we use the attention input of the \emph{previous} Transformer layer
to speculate and prefetch the keys and values of the important tokens for the
\emph{current} layer. The speculation is done by performing a minimal rehearsal
of attention computation of the current layer in the preceding layer. 
This allows for reducing the waste of PCIe bandwidth by only transferring the
keys and values critical for attention computation while preserving model
accuracy.
In addition, although the KV cache is offloaded to CPU memory, which is much
cheaper and larger than GPU memory, we manage the KV cache pool size so as not
to put too much pressure on CPU memory. 

As shown in~\figref{framework_diagram}, there are two major components in the
\name{} runtime. The first includes the Partial Weight Index Generation
Controller, KV Selection Controller, and Inference Controller. These
controllers cooperate to speculate and prefetch the critical KV cache entries
while serving LLM inference.
Additionally, to aid in prefetching, the Skewing Controller performs offline
modifications on the model weights.
We explain each operation in~\ssecref{prefetching}. 
The second component is the Pool Manager. It manages the KV cache pool on CPU
memory under CPU memory pressure, which we discuss in~\ssecref{management}.

%%%%%%%%%%%%%%%%%%%%%%%%%%%%%%%%%%%%%%%%%%%%%%%%%%%%%%%%%%%%%%%%%%%%%%%%
\putssec{opportunity}{Prefetching Opportunities}
In the following, we first explain why using the attention input of the
previous layer for speculation makes sense. We then show how we modify the
query and key weight matrices to make our speculation far more effective.

%%%%%%%%%%%%%%%%%%%%%%%%%%%%%%%%%%%%%%%%%%%%%%%%%%%%%%%%%%%%%%%%%%%%%%%%
\myparagraph{Attention Input Similarity.}
Our prefetching module builds on the key observation that the attention inputs
of consecutive attention layers are highly similar in LLMs. There are two major
reasons behind this. 
The first is the existence of \emph{outliers} in LLMs, as discussed
in~\ssecref{outlier}, and the second is due to layer normalization (LayerNorm).

To begin with, the input to the Transformer block $i$ (\mm{Tblock\_in}{i}) can
be formulated as follows:
\begin{equation} 
  \small 
  \begin{aligned} 
    Attn\_out_{i-1}\, &= \, Attn(LN(Tblock\_in_{i-1}))  \\
    FFN\_out_{i-1} \, &= \, FFN(LN(Tblock\_in_{i-1} + Attn\_out_{i-1}))  \\
    Tblock\_in_{i} \, &= \, Tblock\_in_{i-1} + Attn\_out_{i-1} + FFN\_out_{i-1},
    \label{eqn:tblock_input} 
  \end{aligned} 
\end{equation}

\noindent where \mm{Tblock\_in}{i-1} is an input for Layer $i-1$, which is
first layer-normalized ($LN$) and is fed into the attention layer in the
Transformer block.
After performing attention, we obtain the output (\mm{Attn\_out}{i-1}), which
is added to \mm{Tblock\_in}{i-1} because of the residual connection. 
Then, the sum of \mm{Tblock\_in}{i-1} and \mm{Attn\_out}{i-1} is again
layer-normalized and is fed into the FFN layer.
Afterward, we obtain the FFN output (\mm{FFN\_out}{i-1}), which is added to the
sum of \mm{Tblock\_in}{i-1} and \mm{Attn\_out}{i-1} again due to the residual
connection.
Finally, the sum of \mm{Tblock\_in}{i-1}, \mm{FFN\_out}{i-1}, and
\mm{Attn\_out}{i-1} is used as input to the next Transformer block
(\mm{Tblock\_in}{i}).

\begin{figure}[t]
  \includegraphics[width=\linewidth]{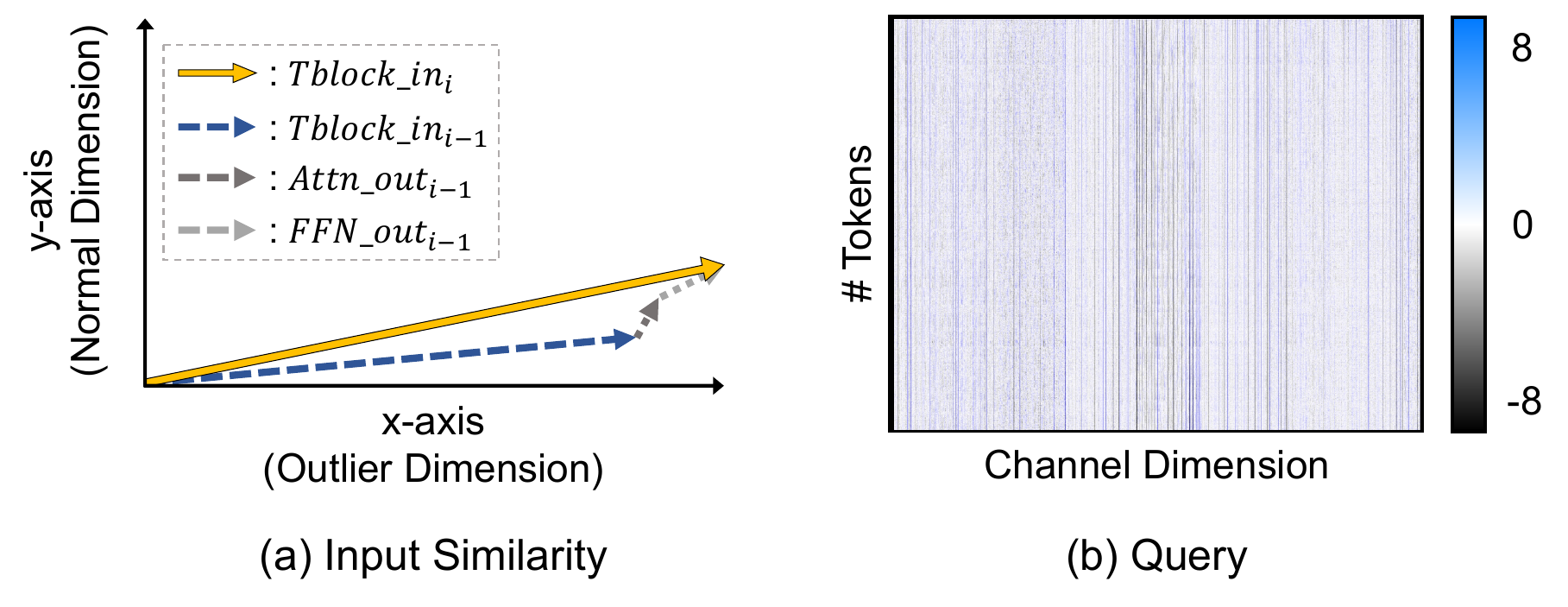}
  \caption{
    (a) Visualization of input similarity between consecutive Transformer
    blocks. (b) Query matrix of Layer 18 of the OPT-13B model. We only show
    channels from 3000 to 4000 for a clearer view of column-wise patterns.
  }
  \vspace{-0.15in}
  \label{fig:input_sim}
\end{figure}

Now, we show why the attention input of Layer $i$ is similar to the one of
Layer $i-1$ with the example in~\figref{input_sim}(a).
In the figure, there are four vectors, each of which corresponds to a term
in~\eqnref{tblock_input}.
The x-axis represents an outlier channel among the model dimension, while the
y-axis represents a normal channel (i.e., other than the outlier channel).
In practice, there exist more normal channels and only a few outlier channels
in the input tensors, but we only present one channel each for both outlier and
normal channels for clarity.

\mm{Tblock\_in}{i-1} is highly skewed along the outlier channel (x-axis) due to
a few outlier channels containing significantly large values compared to those
in the normal channels. 
In contrast, \mm{Attn\_out}{i-1} and \mm{FFN\_out}{i-1} have relatively small
values for both outlier and normal channels (i.e., short vectors).
This is because the attention and FFN inputs are \emph{layer-normalized},
reducing the magnitude of each value.
The small magnitude of the attention and FFN inputs naturally results in their
output values being relatively small compared to \mm{Tblock\_in}{i-1}. 
Consequently, \mm{Tblock\_in}{i} is highly influenced by \mm{Tblock\_in}{i-1},
rather than \mm{Attn\_out}{i-1} or \mm{FFN\_out}{i-1}.  
Highly similar inputs between consecutive Transformer blocks lead to similar
inputs across the attention layers, as the attention input is a
layer-normalized one of the Transformer block input. 

\begin{table}[t]
  \caption{
    Average cosine similarity between the Transformer block input of Layer
    $i$ (\mm{Tblock\_in}{i}) and the other three tensors
    (\mm{Tblock\_in}{i-1}, \mm{Attn\_out}{i-1}, \mm{FFN\_out}{i-1}) across
    the layers. We use a random sentence with 2000 tokens from the PG-19
    dataset~\cite{rae:pot20}.
  }
  \centering
  \resizebox{\linewidth}{!}{%
    {\Large
      \begin{tabular}{c|ccccc}
        \toprule
            Tensors                &\large{OPT-6.7B} &\large{OPT-13B} &\large{OPT-30B} &\large{Llama-2-7B} &\large{Llama-2-13B}  \\
        \midrule
            \mm{Tblock\_in}{i-1}   &\textbf{0.95}    &\textbf{0.96}   &\textbf{0.97}   &\textbf{0.89}      &\textbf{0.91}        \\
            \mm{Attn\_out}{i-1}    &0.29             &0.28            &0.36            &0.31               &0.27                 \\
            \mm{FFN\_out}{i-1}     &0.34             &0.28            &0.35            &0.37               &0.34                 \\
        \bottomrule
      \end{tabular}
    }
  }
  \label{tab:input_similarity}
  \vspace{-0.15in}
\end{table}
\tabref{input_similarity} shows the cosine similarity between
\mm{Tblock\_in}{i} and the other three tensors (\mm{Tblock\_in}{i-1},
\mm{Attn\_out}{i-1}, \mm{FFN\_out}{i-1}).
As shown in the table, \mm{Tblock\_in}{i} is highly dependent on the
\mm{Tblock\_in}{i-1} rather than others. 
\name{} leverages this key observation to speculate on the attention pattern of
Layer $i$ using the attention input of Layer $i-1$.
Note that \mm{Tblock\_in}{} gradually changes across the layers; the inputs to
distant layers are distinct.

%%%%%%%%%%%%%%%%%%%%%%%%%%%%%%%%%%%%%%%%%%%%%%%%%%%%%%%%%%%%%%%%%%%%%%%%
\myparagraph{Skewed Partial Weight.}
We observe that the attention score highly depends on a few columns in the
query and key matrices.
\figref{input_sim}(b) shows the values in a query matrix of Layer 18 of the
OPT-13B model, where the column-wise patterns indicate that there exist certain
columns with large magnitudes in the matrix; we observe the same patterns in
the key and query matrices across different layers and models.
The large magnitude columns have a great impact on the attention pattern
because the dot product between the query and key is highly affected by these
few columns. 
The column-wise pattern in the attention input indicates that there is little
variance between each row in the outlier channels. Thus, the dot product
between any row of the attention input and a column of the weight matrix could
have a similarly large magnitude, which induces the outlier channels in the
query and key matrices.

Going one step further, if we make a few columns in the query and key matrices
have much larger magnitude than others, a much smaller number of columns
significantly affects the attention pattern. 
We can do this by multiplying the query and key weight matrices with the same
orthogonal matrix $A$. Since the transpose of the orthogonal matrix is the
inverse of itself, the proposed operation does not change the final result, as
shown in~\eqnref{compute-decomposition0} (i.e., this is mathematically
equivalent to $QK^{T}$, not an approximation):
\begin{equation}
  \small
  \begin{aligned}
    \tilde{Q} &= X_{a} \times W_{Q} \times A, \quad\quad\quad \tilde{K} = X_{a} \times W_{K} \times A          \\ 
    \tilde{Q} \times \tilde{K}^{T} &= X_{a} \times W_{Q} \times A \times (X_{a} \times W_{K} \times A)^T       \\
                                   &= X_{a} \times W_{Q} \times A \times A^{T} \times W_{K}^{T} \times X_{a}^T \\
                                   &= X_{a} \times W_{Q} \times W_{K}^{T} \times X_{a}^T                       \\
                                   &= X_{a} \times W_{Q} \times (X_{a} \times W_{K})^{T}                       \\
                                   &= Q \times K^{T},
  \label{eqn:compute-decomposition0}
  \end{aligned}
\end{equation}
where $\tilde{Q}$ and $\tilde{K}$ are skewed query and key matrices, while
$W_{Q}$ and $W_{K}$ are query and key weight matrices. 
$X_{a}$ denotes the attention input.
We set the orthogonal matrix $A$ whose direction aligns with the direction that
the query matrix stretches the most. Specifically, we first decompose the query
matrix using SVD and obtain $\mathbf{U}$, $\Sigma$, and $\mathbf{V}$. We then
set $A$ to orthogonal matrix $\mathbf{V}$ to align the column vectors with the
standard unit vectors as $\mathbf{V}^{T}\mathbf{A}=\mathbf{V}^{T}\mathbf{V}=I$,
where $I$ is an identity matrix. We formulate the skewed query matrix as
follows:
\begin{equation}
  \small
  \begin{aligned}
    \tilde{Q} =& \,\, Q \times A = \mathbf{U} \Sigma \mathbf{V}^{T} \times A = \mathbf{U} \Sigma \mathbf{V}^{T} \times \mathbf{V} 
  \label{eqn:compute-decomposition}
  \end{aligned}
\end{equation}
In this way, we can make a few columns with large magnitudes in $\tilde{Q}$
without altering the result of computation, as discussed in~\ssecref{svd}. 

\begin{figure}[t]
  \includegraphics[trim=0.0in 0.0in 0.0in 0.30in,clip,width=\linewidth]{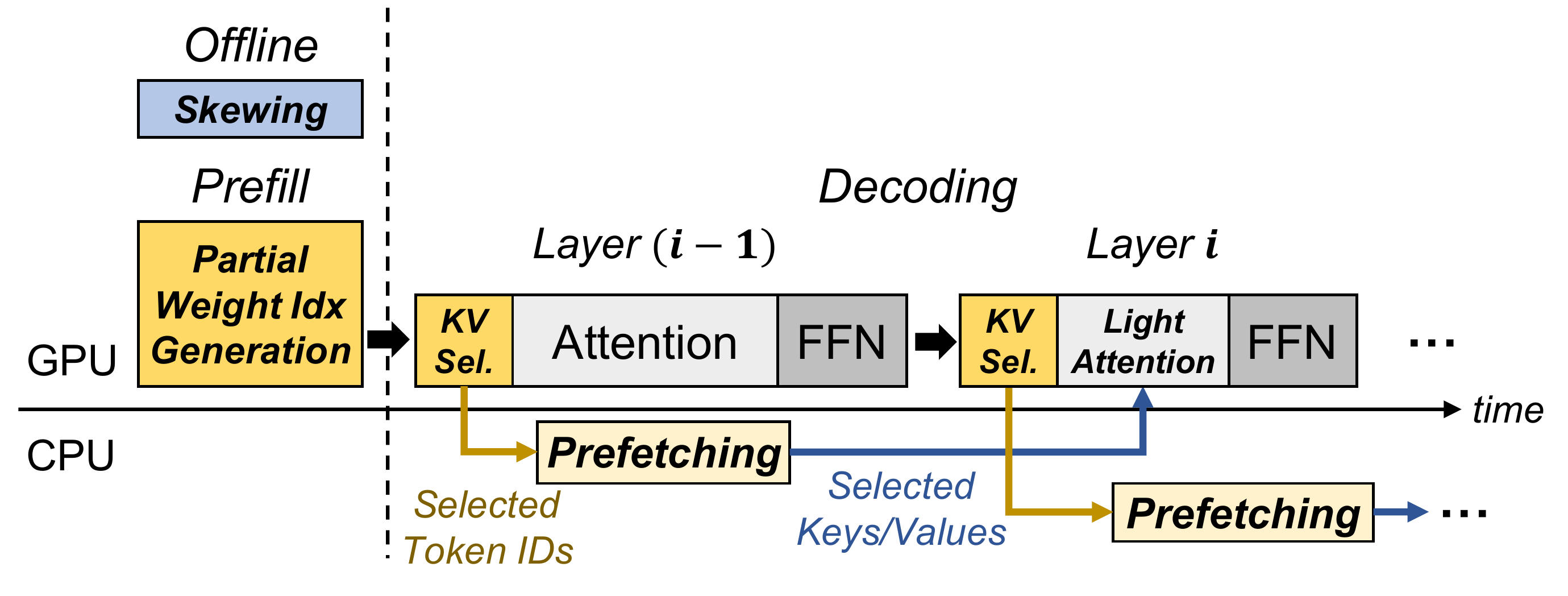}
  \caption{Operation flow of the prefetching module of \name{}.}
  \vspace{-0.15in}
  \label{fig:prefetch_overview}
\end{figure}

%%%%%%%%%%%%%%%%%%%%%%%%%%%%%%%%%%%%%%%%%%%%%%%%%%%%%%%%%%%%%%%%%%%%%%%%
\putssec{prefetching}{Efficiently Prefetching KV Cache}
\noindent\textbf{Prefetching Scheme.}
\figref{prefetch_overview} shows the operation flow of the prefetching module
in \name{}. 
In the offline phase, \name{} modifies the \emph{weight} matrices to generate
skewed query and key matrices. To achieve this, \name{} first runs the forward
pass of the model once with a sample input. During this process, \name{}
gathers the query matrix from each layer and performs singular value
decomposition (SVD) of each query matrix.
The skewing matrix ($A_i$) of each layer is obtained using the decomposed
matrices of the query matrix, as shown in~\eqnref{compute-decomposition}. This
matrix is then multiplied with each of the query and key weight matrices in the
corresponding layer. 
Importantly, after the multiplication, the dimensions of the weight matrices
remain unchanged. Note that the skewing is a \emph{one-time} offline process
and does not incur any runtime overhead because we modify the weight matrices
that are invariant at runtime.
As we exploit the column-wise pattern, which stems from the intrinsic property
of the model rather than the input, whenever we compute the query and key for
different inputs after the skewing, the values exhibit a high degree of
skewness, thereby improving the effectiveness of our prefetching module.
Note that skewing does not alter the original functionality. Even with the
skewing, the attention layer produces identical computation results.

\begin{figure}[t]
  \includegraphics[width=\linewidth]{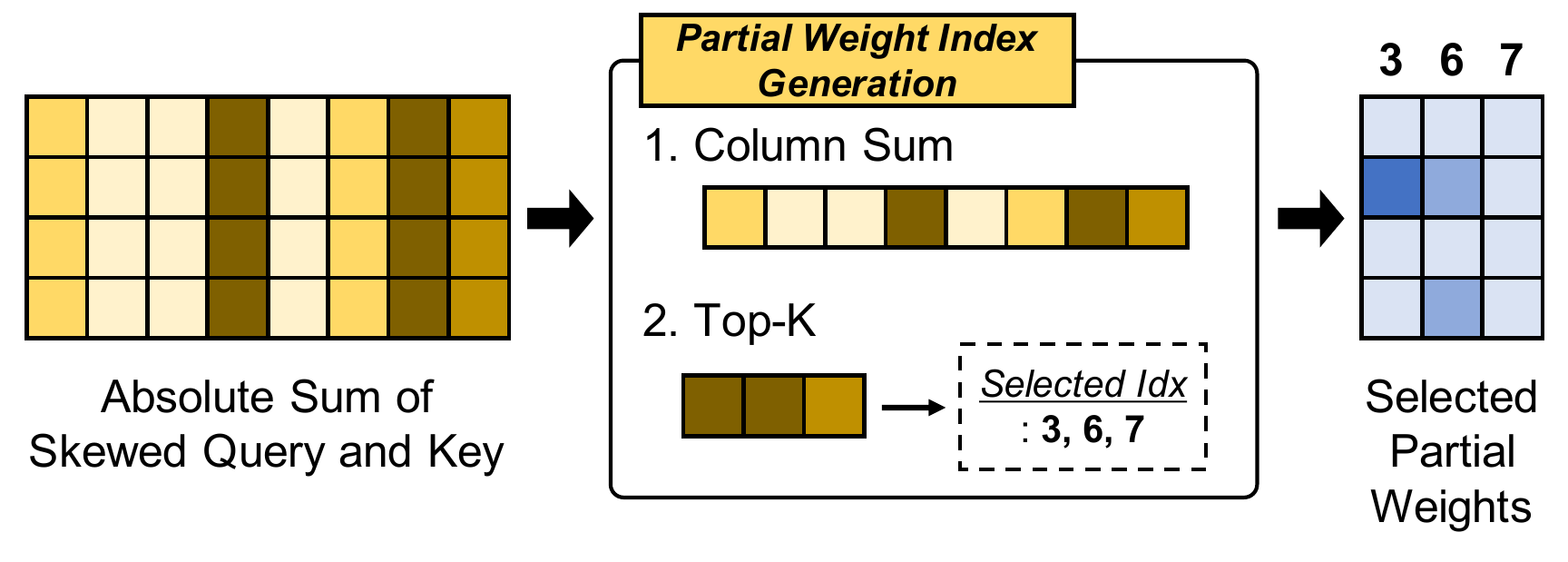}
  \caption{Partial weight generation in the prefill stage.}
  \label{fig:prefill}
\end{figure}

%%%%%%%%%%%%%%%%%%%%%%%%%%%%%%%%%%%%%%%%%%%%%%%%%%%%%%%%%%%%%%%%%%%%%%%%
\myparagraph{Prefill Stage.}
In the prefill stage, \name{} selects several important columns from the query
weight matrix and the key cache to speculate on the attention pattern, and
generates \emph{partial} query weight and key cache matrices used in the
decoding stage.
\figref{prefill} shows how \name{} creates these partial matrices.
Because we multiply each column in the query matrix with the corresponding row
in the transposed key matrix, it is essential to select the \emph{same} column
indices in the query weight matrix and the key cache to obtain a proper
approximation of the attention score.
However, the indices of the outlier columns of the skewed query (${\tilde{Q}}$)
and key (${\tilde{K}}$) matrices may not align exactly.
To obtain partial matrices that capture the outliers, we first take the
element-wise absolute values of the skewed query and key matrices, then add
these two matrices together.
This helps us calculate the sum of each column and perform top-$k$ operation
only once while accommodating the outlier columns of both query and key
matrices.
We then sum the elements in each column and select the top-$k$ columns in the
matrix; we choose 30\% of the columns in our work.
Using the sum of column values captures the global trend of each column while
minimizing the effect of variance in each row.
The selected columns better approximate the attention pattern because of the
use of skewed query and key matrices.

\begin{figure}[t]
  \includegraphics[width=\linewidth]{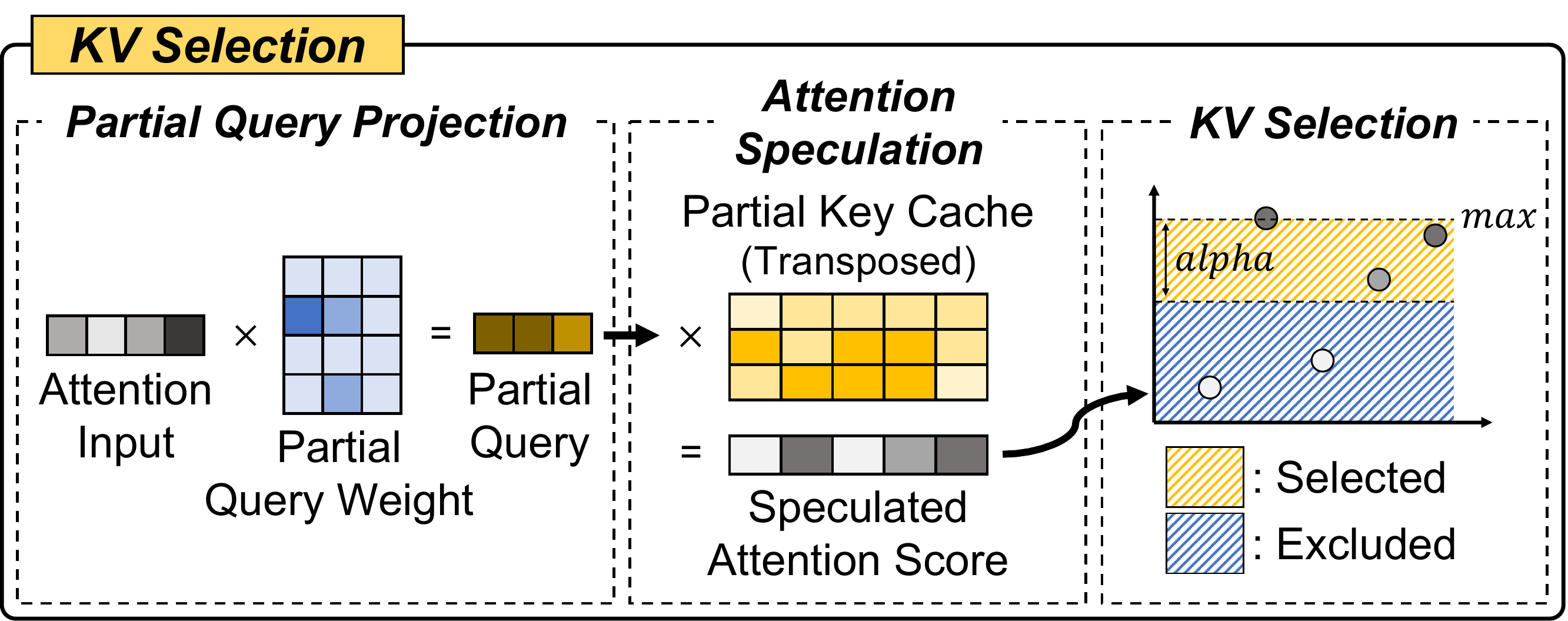}
  \caption{Attention score speculation in the decoding stage.}
  \vspace{-0.15in}
  \label{fig:decode}
\end{figure}

%%%%%%%%%%%%%%%%%%%%%%%%%%%%%%%%%%%%%%%%%%%%%%%%%%%%%%%%%%%%%%%%%%%%%%%%
\myparagraph{Decoding Stage.}
In the decoding stage, \name{} speculates on the attention pattern of the next
layer and determines the critical keys and values to prefetch. 
\figref{decode} shows how \name{} computes the \emph{speculated} attention
score.
At Layer $i-1$, we use the partial query weight matrix and key cache of Layer
$i$, which are identified in the prefill stage, along with the attention input
of Layer $i-1$.
After multiplying the partial query and partial key cache, \name{} selects
tokens with high attention scores. 

We set the threshold considering the maximum value of the speculated attention
score.
We select only the tokens with an attention score greater than the maximum
score subtracted by \textit{alpha}.  
It is noted that subtraction from the attention score results in division after
softmax.  
For example, assume that the attention score of the 3$^{\mathrm{rd}}$ token is
the maximum attention score minus 5. Once we apply softmax to the attention
scores, the attention weight of the 3$^{\mathrm{rd}}$ token is the maximum
attention weight divided by $e^5\approx148.4$. 
Even though we do not use this token, it does not noticeably hurt the accuracy
of the model since it accounts for less than 1\% of importance
($\approx1/148.4$) after softmax.  
Thus, \name{} only prefetches the keys and values of the tokens with an
attention score larger than the highest attention score minus \textit{alpha}.
As multiple attention heads are computed in parallel, we ensure that each head
in the same layer fetches the same number of tokens by averaging the number of
tokens between the maximum score and the threshold across the heads.

By reducing the amount of KV cache to load and compute, \name{} effectively
reduces the loading latency (i.e., data transfer from CPU to GPU) while
maintaining an output quality similar to that of the original model with a full
KV cache.
Moreover, as \name{} does not require a fixed number of tokens to load from CPU
memory, it utilizes only the necessary PCI interconnect bandwidth.
\name{} initiates speculation and prefetching from Layer 1 because the
outliers, which are essential for exploiting input similarity, emerge during
the computation in Layer 0.

\begin{figure*}[ht]
  \includegraphics[clip,trim=0in 0.15in 0in 0in,width=\textwidth]{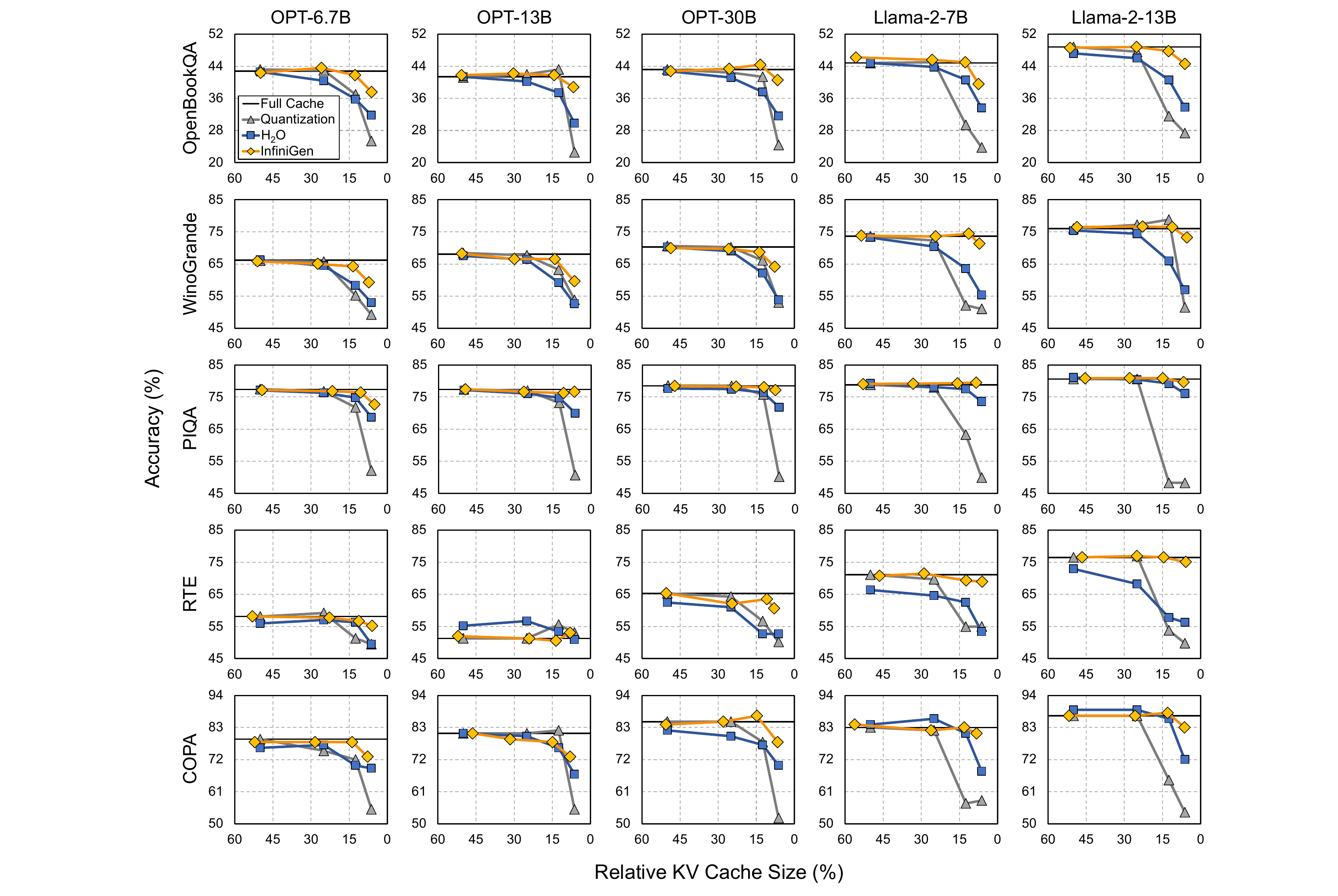}
  \caption{Accuracy of LLMs on 5-shot tasks in lm-evaluation-harness.}
  \vspace{-0.15in}
  \label{fig:lm-eval-harness}
\end{figure*}

%%%%%%%%%%%%%%%%%%%%%%%%%%%%%%%%%%%%%%%%%%%%%%%%%%%%%%%%%%%%%%%%%%%%%%%%
\putssec{management}{KV Cache Pool Management}
We manage the KV cache as a pool, offloading to the CPU memory and prefetching
only the essential amount to the GPU. While CPU memory is more affordable and
larger than GPU memory, it still has limited capacity. Hence, for certain
deployment scenarios, it might be crucial to confine the size of the KV cache
pool and remove less important KV entries that are infrequently selected by
query tokens.
We extend the design to incorporate a user-defined memory size limit. During
runtime, when the size of the CPU memory reaches a user-defined limit, the KV
cache pool manager selects a victim KV entry for eviction. Subsequently, the
manager overwrites the selected victim with the newly generated key and value,
along with updating the corresponding partial key cache residing in the GPU.
It is noted that the order of KV entries can be arbitrary, as long as the key
and value of the same token maintain the same relative location in the KV cache
pool. 

The policy of selecting a victim is important since it directly impacts model
accuracy.
We consider a counter-based policy along with two widely used software cache
eviction policies: FIFO~\cite{ber:ber20,yan:zha23, yan:pol22} and
Least-Recently-Used (LRU)~\cite{memcached}. 
The FIFO-based policy is easy to implement with low overhead but results in a
relatively large accuracy drop since it simply evicts the oldest residing
token.
The LRU-based policy generally exhibits a smaller decrease in accuracy but
often entails a higher runtime overhead. In general, LRU-based policy uses a
doubly linked list with locks to promote accessed objects to the head, which
requires atomic memory updates for accessed KV entries.  
In the case of the counter-based policy, the pool manager increments a counter
for each prefetched KV entry and selects a victim with the smallest count in
the KV cache pool. If any counter becomes saturated, all the counter values are
reduced by half.
We observe that the counter-based policy and the LRU-based one show comparable
model accuracy, which we discuss in \ssecref{accuracy}.
We opt for a counter-based approach due to its simpler design and to avoid
atomic memory updates for better parallelism.

\putsec{eval}{Evaluation}
\putssec{method}{Experimental Setup}
\noindent\textbf{Model and System Configuration.}
We use Open Pre-trained Transformer (OPT) models~\cite{zha:rol22} with 6.7B,
13B, and 30B parameters for evaluation. The 7B and 13B models of
Llama-2~\cite{tou:mar23} are also used to demonstrate that \name{} works
effectively across different model architectures. We run the experiments on a
system equipped with an NVIDIA RTX A6000 GPU~\cite{a6000} with 48GB of memory
and an Intel Xeon Gold 6136 processor with 96GB of DDR4-2666 memory. PCIe 3.0
$\times$16 interconnects the CPU and GPU.

%%%%%%%%%%%%%%%%%%%%%%%%%%%%%%%%%%%%%%%%%%%%%%%%%%%%%%%%%%%%%%%%%%%%%%%%
\myparagraph{Workload.}
We evaluate using few-shot downstream tasks and language modeling datasets.
We use five few-shot tasks from the lm-{evaluation}-harness
benchmark~\cite{gao:tow21}: COPA~\cite{roe:cos11}, OpenBookQA~\cite{mih:cla18},
WinoGrande~\cite{sak:bra21}, PIQA~\cite{yon:row20}, and RTE~\cite{wan:sin18}. 
The language modeling datasets used are WikiText-2~\cite{mer:xio16} and Penn
Treebank (PTB)~\cite{mar:kim94}. Additionally, randomly sampled sentences from
the PG-19 dataset~\cite{rae:pot20} are used to measure the speedup with long
sequence lengths.

%%%%%%%%%%%%%%%%%%%%%%%%%%%%%%%%%%%%%%%%%%%%%%%%%%%%%%%%%%%%%%%%%%%%%%%%
\myparagraph{Baseline.}
We use two inference environments that support KV cache offloading: CUDA
Unified Virtual Memory (UVM)~\cite{tyl:ron21} and FlexGen~\cite{she:zhe23}. On
UVM, all data movements between the CPU and GPU are implicitly managed by the
UVM device driver, thereby enabling offloading without requiring intervention
from the programmer. In contrast, FlexGen uses explicit data transfers between
the CPU and GPU. For the FlexGen baseline, unless otherwise specified, we
explicitly locate all the KV cache in the CPU memory. 
The model parameters are stored in the GPU memory as much as possible, with the
remainder in the CPU memory. 
We compare \name{} with two different KV cache management methods:
\ho{}~\cite{zha:she23} and Quantization~\cite{she:zhe23}. \ho{}, a recent
method in KV cache management, maintains the KV cache of the important or
recent tokens by assessing the importance of each token and discarding others.
Quantization-based compression applies group-wise asymmetric quantization to
the KV cache.

\begin{figure}[t]
  \includegraphics[width=\linewidth]{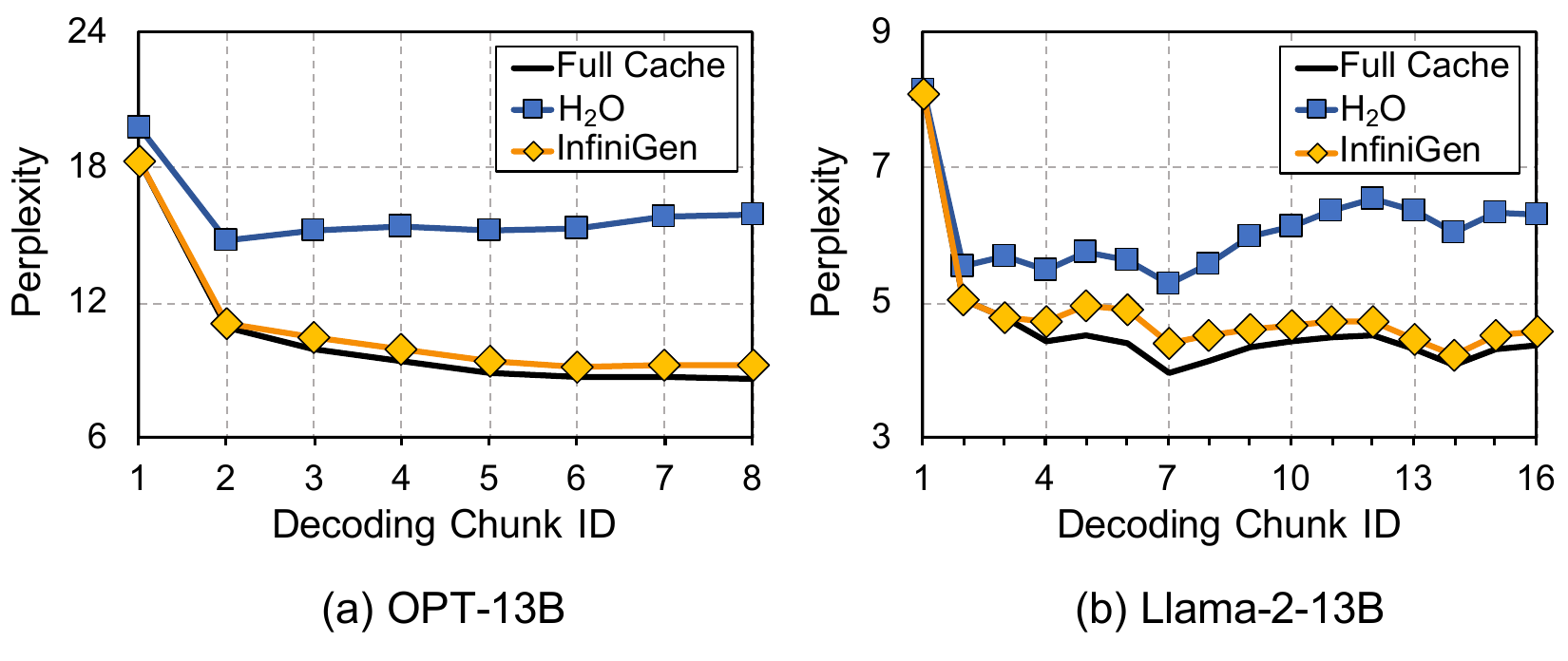}
  \caption{
    Perplexity of OPT-13B and Llama-2-13B for WikiText-2 dataset. Lower is
    better. Perplexity is computed for each decoding chunk that contains 256
    tokens.
  }
  \vspace{-0.15in}
  \label{fig:seqlen_acc}
\end{figure}

%%%%%%%%%%%%%%%%%%%%%%%%%%%%%%%%%%%%%%%%%%%%%%%%%%%%%%%%%%%%%%%%%%%%%%%%
\myparagraph{Key Metric.}
We evaluate accuracy (\%) to assess the impact of approximation when \name{},
\ho{}, and Quantization are used. For the language modeling tasks with
WikiText-2 and PTB, we use perplexity as a metric; lower perplexity means
better accuracy. To present performance improvements, we measure the wall clock
time during inference with varying batch sizes and sequence lengths.
The partial weight ratio is set to 0.3. 
We set alpha to 4 for OPT and 5 for Llama-2, resulting in using less than 10\%
of the KV cache on average across the layers.
For each layer, we allow sending up to 20\% of the total KV cache to the GPU if
it contains more candidates.
The partial weight ratio and alpha are determined based on a sensitivity study
for each model to balance accuracy and inference latency, which we discuss
in~\ssecref{sensitivity}.

%%%%%%%%%%%%%%%%%%%%%%%%%%%%%%%%%%%%%%%%%%%%%%%%%%%%%%%%%%%%%%%%%%%%%%%%
\putssec{accuracy}{Language Modeling}
\noindent\textbf{Accuracy on lm-evaluation-harness.}
\figref{lm-eval-harness} shows the accuracy of the baselines and \name{} across
different models with 5-shot tasks. The relative KV cache size indicates the
size of the KV cache involved in the attention computation compared to the
full-cache baseline (e.g., a 10\% relative KV cache size means that 10\% of the
full KV cache size is used).
\name{} consistently shows better accuracy across the models and tasks when the
relative KV cache size is less than 10\%, whereas the others exhibit a
noticeable accuracy drop due to insufficient bit widths (Quantization) or
permanent KV cache elimination (\ho{}). This implies that our proposed solution
can effectively reduce the KV cache transfer overhead while preserving model
accuracy.
For relative KV cache sizes larger than 10\%, the accuracy with \name{} closely
matches that of the full-cache baseline. In some cases, \name{} even shows
slightly better accuracy than the full-cache baseline.
This is likely because reducing the amount of the KV cache participating in the
attention computation can help the model focus more on critical tokens.

%%%%%%%%%%%%%%%%%%%%%%%%%%%%%%%%%%%%%%%%%%%%%%%%%%%%%%%%%%%%%%%%%%%%%%%%
\myparagraph{Sequence Length.}
\figref{seqlen_acc} shows the perplexity of two different models with \name{}
and the baselines, as the sequence length increases.
In this experiment, \ho{} is configured to use the same amount of KV cache as
\name{}. The sequence lengths are 2048 and 4096 for OPT-13B and Llama-2-13B,
respectively.
For a clearer view, we evaluate perplexity with consecutive 256 tokens as a
group, which is referred to as a decoding chunk in the figure.
The results show that even though the sequence length becomes longer (i.e., the
decoding chunk ID increases), the perplexity of \name{} remains consistently
comparable to the full-cache baseline, while \ho{} shows an increasing
divergence from the baseline. 
\ho{} suffers from permanent KV cache elimination and may not retain a
sufficient amount of KV cache in certain layers due to its fixed budget. In
contrast, \name{} dynamically computes attention using only the essential
amount of KV cache for each layer. 
The difference is likely to widen as the models become capable of handling much
longer sequences.

\begin{table}[t]
  \caption{
    Perplexity on WikiText-2 and PTB with 2048 sequence length with or without
    KV cache memory limits. Lower is better.
  }
  \centering
  \huge
  \resizebox{\linewidth}{!}{%
    \renewcommand{\arraystretch}{1.3}
    \begin{tabular}{c|rr|rr|rr|rr|rr}
      \toprule[2pt]
      \multirow{2}{*}{Scheme}  &\multicolumn{2}{c}{OPT-6.7B} &\multicolumn{2}{c}{OPT-13B} &\multicolumn{2}{c}{OPT-30B} &\multicolumn{2}{c}{Llama-2-7B} &\multicolumn{2}{c}{Llama-2-13B}  \\
      \cmidrule(lr{1em}){2-11}
      &\multicolumn{1}{c}{Wiki} &\multicolumn{1}{c}{PTB} &\multicolumn{1}{c}{Wiki} &\multicolumn{1}{c}{PTB} &\multicolumn{1}{c}{Wiki} &\multicolumn{1}{c}{PTB} &\multicolumn{1}{c}{Wiki} &\multicolumn{1}{c}{PTB} &\multicolumn{1}{c}{Wiki} &\multicolumn{1}{c}{PTB}  \\
      \midrule{}
                {100\%}                   & 11.68  & 13.86  & 10.55  & 12.78  &10.14  & 12.31  & 5.69  & 22.53 &  5.25 & 31.94  \\
      \midrule{}                                                                                 
                {80-FIFO\%}               & 19.64  & 16.82  & 30.99  & 33.84  & 30.66 & 35.45  & 22.26 & 61.88 & 21.41 & 32.34  \\
                {80-LRU\%}                & 11.68  & 13.85  & 10.55  & 12.78  & 10.14 & 12.31  & 5.69  & 22.53 & 5.25  & 31.94  \\
                {80-Counter\%}            & 11.68  & 13.86  & 10.55  & 12.78  & 10.14 & 12.31  & 5.69  & 22.53 & 5.25  & 31.94  \\
      \bottomrule[2pt]
    \end{tabular}
  }
  \label{tab:compression_ppl}
  \vspace{-0.10in}
\end{table}

\begin{figure}[t]
  \includegraphics[width=\linewidth]{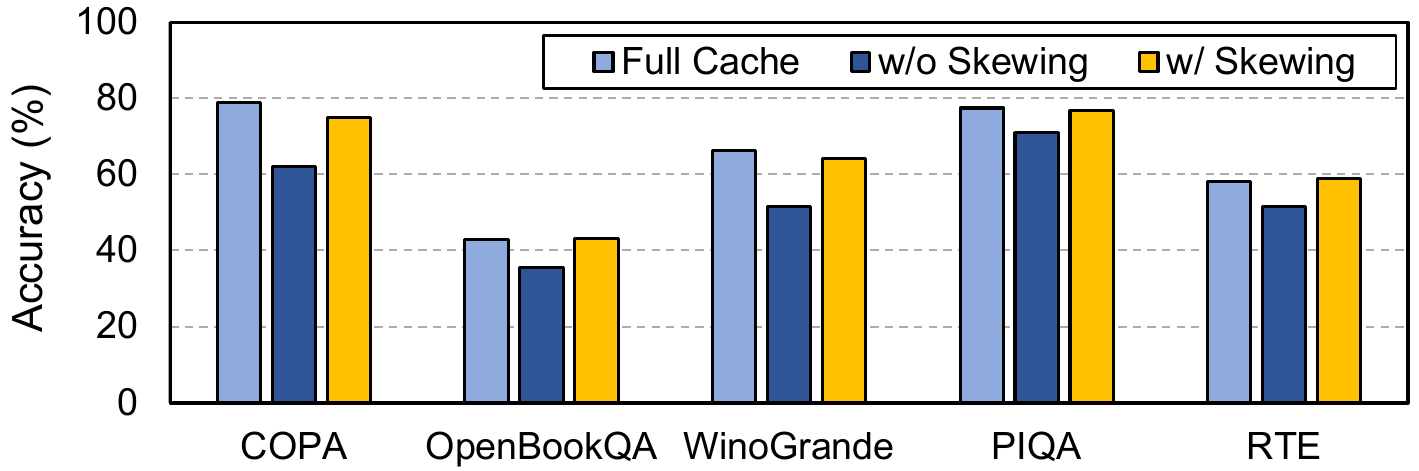}
  \caption{
    Accuracy on the lm-evaluation-harness benchmark with or without skewing on
    OPT-6.7B.
  }
  \vspace{-0.15in}
  \label{fig:skewing}
\end{figure}

%%%%%%%%%%%%%%%%%%%%%%%%%%%%%%%%%%%%%%%%%%%%%%%%%%%%%%%%%%%%%%%%%%%%%%%%
\myparagraph{Effect of Skewing.}
\figref{skewing} shows the accuracy with or without key/query skewing on the
OPT-6.7B model. For the experiment, we use a fixed KV cache budget of 20\%,
instead of using a dynamic approach, to clearly show the effect of skewing. We
observe that several language models (e.g., Llama-2) show a small drop in
accuracy without skewing. For some models such as OPT-6.7B, however, we see a
large accuracy drop if we do not apply the skewing method as shown
in~\figref{skewing}. 
This indicates that in the case of OPT-6.7B, the partial weight does not
adequately represent the original matrix without skewing. After applying our
skewing method, we achieve accuracy similar to the full-cache baseline.  Our
skewing method effectively skews key and query matrices such that a few columns
can better represent the original matrices.

%%%%%%%%%%%%%%%%%%%%%%%%%%%%%%%%%%%%%%%%%%%%%%%%%%%%%%%%%%%%%%%%%%%%%%%%
\myparagraph{KV Cache Pool Management.}
\tabref{compression_ppl} shows the perplexity of five different models with or
without limiting the memory capacity for WikiText-2 and PTB. We compare
FIFO-based, LRU-based, and Counter-based victim selection policies in
\ssecref{management} under the 80\% memory limit of a full KV cache. We also
present the perplexity results with no memory limit (100\%). The FIFO-based
approach shows the worst model performance because it simply deletes the oldest
KV entry regardless of its importance. The LRU and Counter-based approaches
show perplexity that is almost similar to that with no memory limit. We choose
a Counter-based victim selection policy instead of an LRU-based approach
because the LRU-based approach typically needs to maintain a doubly linked list
queue with locks for atomic memory updates. 

\begin{figure}[t]
  \includegraphics[width=\linewidth]{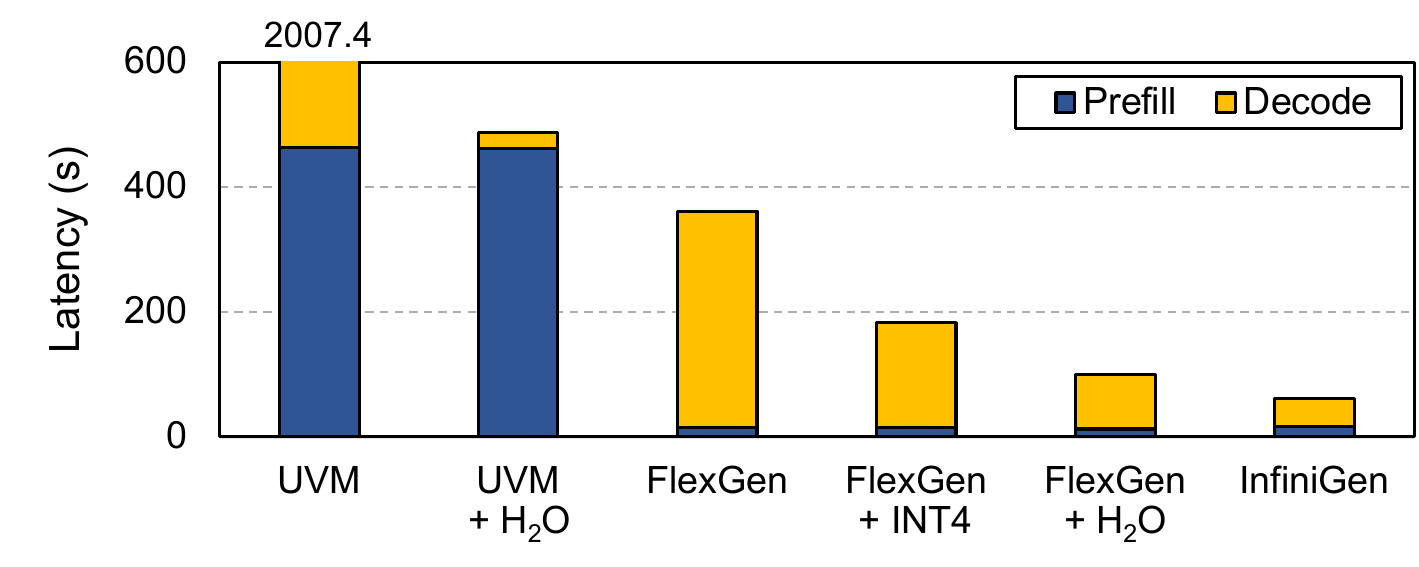}
  \caption{
    Inference latency on OPT-13B with a sequence length of 2048 (1920 input and
    128 output tokens) and a batch size of 20.
  }
  \vspace{-0.15in}
  \label{fig:e2e_latency}
\end{figure}

%%%%%%%%%%%%%%%%%%%%%%%%%%%%%%%%%%%%%%%%%%%%%%%%%%%%%%%%%%%%%%%%%%%%%%%%
\putssec{speedup}{Performance}
In this section, we refer to \ho{} (with a KV cache budget of 20\%) and 4-bit
quantization implemented on top of FlexGen as \ho{} and INT4. 

%%%%%%%%%%%%%%%%%%%%%%%%%%%%%%%%%%%%%%%%%%%%%%%%%%%%%%%%%%%%%%%%%%%%%%%%
\myparagraph{Inference Latency.}
\figref{e2e_latency} shows the inference latency including the prefill and
decoding stages.
We use the OPT-13B model with 1920 input tokens, 128 output tokens, and a batch
size of 20. 
\name{} achieves 1.63$\times$-32.93$\times$ speedups over the baselines. The
performance benefit mainly comes from the significantly reduced amount of KV
cache to load from the CPU memory due to our dynamic approach. 

UVM shows an extremely long latency because the working set size (i.e., the
size of the model parameters and KV cache) is larger than the GPU memory
capacity, thereby leading to frequent page faults and data transfers between
the CPU and GPU.
The prefill stage of UVM + \ho{} also shows a long latency due to the page
faults and data transfers. However, because all required data are migrated to
the GPU memory after the prefill stage, UVM + \ho{} shows a substantially
shorter decoding latency. 
FlexGen loads the full KV cache with high precision (i.e., FP16) from the CPU
memory for every attention computation. On the other hand, INT4 and \ho{} load
relatively small amounts of the data from the CPU because of the low-bit data
format (INT4) or a smaller size of the KV cache (\ho{}). However, they still
load larger amounts of data than \name{}; even with low precision, INT4 loads
the KV cache of all the previous tokens; \ho{} always loads the same amount of
data no matter how many tokens are actually important in each layer. 
As a result, \name{} achieves better performance than both of them.

\begin{figure}[t]
  \includegraphics[clip,trim=0in 0.15in 0in 0in, width=\linewidth]{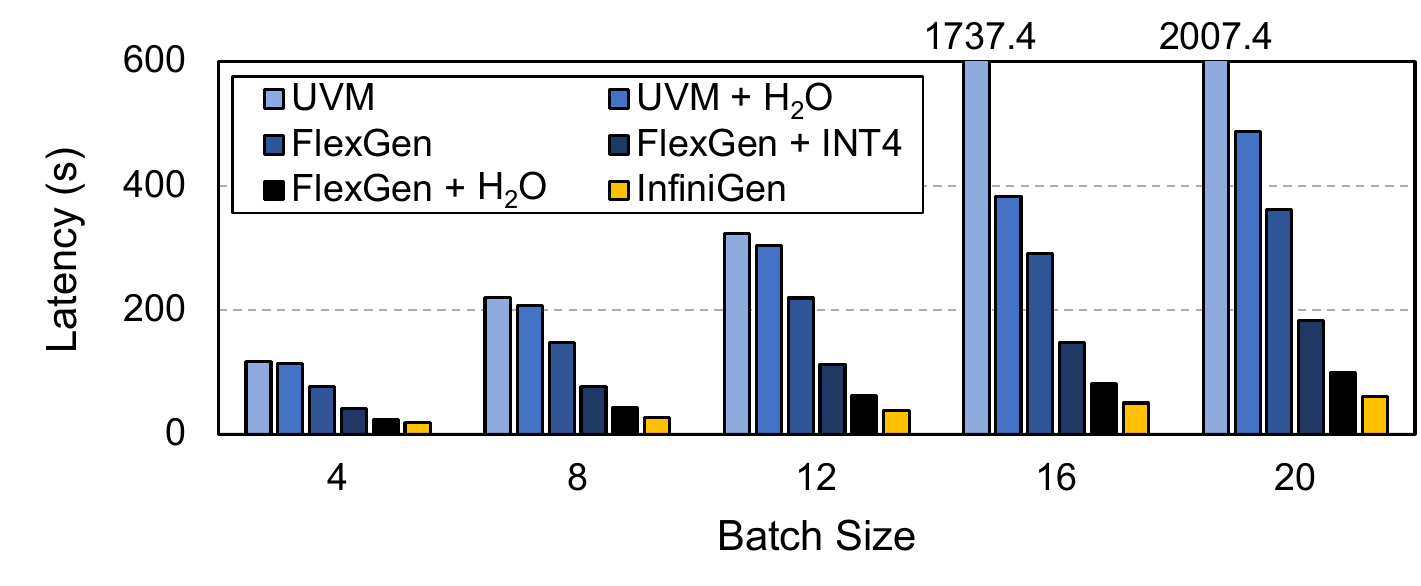}
  \caption{
    Inference latency for 5 different batch sizes on OPT-13B with a sequence
    length of 2048 (1920 input and 128 output tokens).
  }
  \vspace{-0.15in}
  \label{fig:batch_size}
\end{figure}

%%%%%%%%%%%%%%%%%%%%%%%%%%%%%%%%%%%%%%%%%%%%%%%%%%%%%%%%%%%%%%%%%%%%%%%%
\myparagraph{Batch Size.}
\figref{batch_size} shows the inference latency across different batch sizes.
The results show that \name{} achieves lower latency than others across the
batch sizes (1.28$\times$-34.64$\times$). As the batch size increases, the
performance gap between \name{} and others becomes larger. UVM and UVM + \ho{}
show increasing latency mainly due to frequent page faults in the prefill
stage. 
For UVM, the latency also rapidly increases at a batch size of 16 because the
working set size exceeds the GPU memory capacity for \emph{both} prefill and
decoding stages. As the batch size keeps increasing, UVM + \ho{} will face the
same problem as well.

The latency of FlexGen almost linearly increases with the batch size because
the KV cache transfer occupies the majority of the inference latency. As we
increase the batch size from 4 to 20, the throughput (tokens per second) of
\name{} increases from 27.36 to 41.99, while INT4 and \ho{} offer a small
increase in throughput (from 12.22 to 14.02 and from 21.31 to 25.70 each).
By dynamically adjusting the amount of the KV cache to load, \name{} achieves
scalable performance across the batch sizes.

\begin{figure}[t]
  \includegraphics[width=\linewidth]{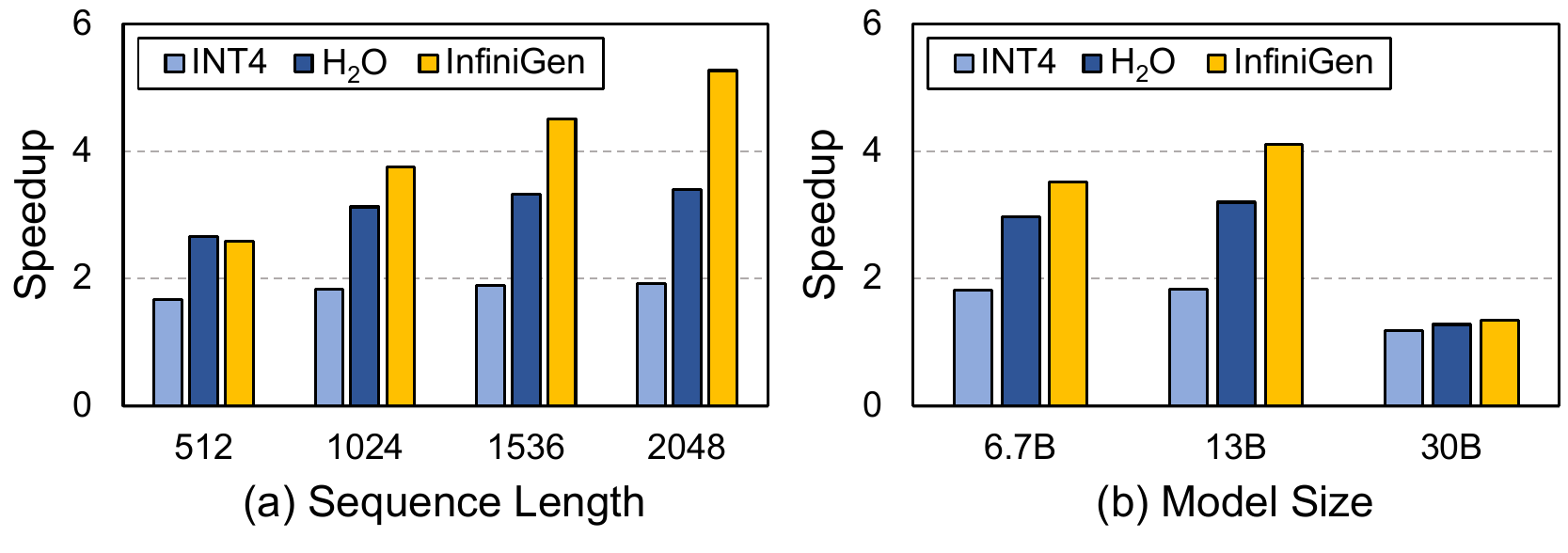}
  \caption{
    Speedup over the FlexGen baseline across (a) sequence lengths and
    (b) model sizes.
  }
  \vspace{-0.15in}
  \label{fig:seq_len}
\end{figure}

%%%%%%%%%%%%%%%%%%%%%%%%%%%%%%%%%%%%%%%%%%%%%%%%%%%%%%%%%%%%%%%%%%%%%%%%
\myparagraph{Sequence Length.}
\figref{seq_len}(a) shows the speedup of INT4, \ho{}, and \name{} over FlexGen
on OPT-13B across different sequence lengths.
With a batch size of 8, we use four different input/output configurations.
Each configuration comprises 128 output tokens and 384, 896, 1408, 1920 input
tokens (i.e., a total number of tokens ranging from 512 to 2048).
The speedup of \name{} continues to increase across the sequence lengths (up to
5.28$\times$), whereas INT4 and \ho{} show saturating speedups (up to
1.92$\times$ and 3.40$\times$). 
This suggests that neither INT4 nor \ho{} provides a scalable solution for KV
cache management. 
INT4 shows a negligible increase in speedup due to the inherent growth in the
size of the KV cache.
Similarly, \ho{} lacks scalability due to its fixed ratio of the KV cache
budget; as the sequence length increases, \ho{} stores and loads more KV cache.

Even though the sequence length increases, the number of tokens that each token
attends to does not increase linearly. For instance, in the OPT-13B model, we
count the number of important tokens with attention scores larger than
$(max-4)$ and identify that, on average, 37, 60, 66, and 73 tokens are assessed
as \emph{important} for sequence lengths of 512, 1024, 1536, and 2048,
respectively.
\ho{}, which employs 20\% of a fixed KV cache budget, loads 409 tokens for the
sequence length of 2048, while only 73 tokens are relatively important. 
In contrast, \name{} naturally captures this trend (i.e., a non-linear increase
in the number of important tokens) by dynamically observing the speculated
attention score.

%%%%%%%%%%%%%%%%%%%%%%%%%%%%%%%%%%%%%%%%%%%%%%%%%%%%%%%%%%%%%%%%%%%%%%%%
\myparagraph{Model Size.}
\figref{seq_len}(b) shows the speedup of INT4, \ho{}, and \name{} over FlexGen
on three different model sizes. We use 1920 input tokens and 128 output tokens
with a batch size of 4 for the experiment. 
The results show that \name{} outperforms others across the model sizes. As the
model size increases from 6.7B to 13B, the speedup of \name{} also increases by
1.17$\times$, while others do not lead to a noticeable increase in speedup.
For most of the layers, \name{} loads a smaller amount of KV cache than \ho{}
because a relatively small number of tokens are needed. Thus, \name{} performs
better than \ho{} as the model size becomes larger due to the increased number
of Transformer blocks.
For the 30B model, the model parameters do not fit in the GPU memory. 
As such, we offload 30\% of the model parameters to the CPU. In this case, the
size of the offloaded parameters is 1.7$\times$ larger than the KV cache size.
Even so, \name{} shows a 1.34$\times$ speedup over FlexGen, while others
achieve 1.18$\times$ and 1.28$\times$ each. 

\putsec{discussion}{Analysis and Discussion}
\putssec{sensitivity}{Sensitivity Study}
We use the OPT-6.7B model with 1920 input tokens, 128 output tokens, and a
batch size of 8. The accuracy is evaluated with the WinoGrande task in
lm-evaluation-harness.

\begin{figure}[t]
  \includegraphics[width=\linewidth]{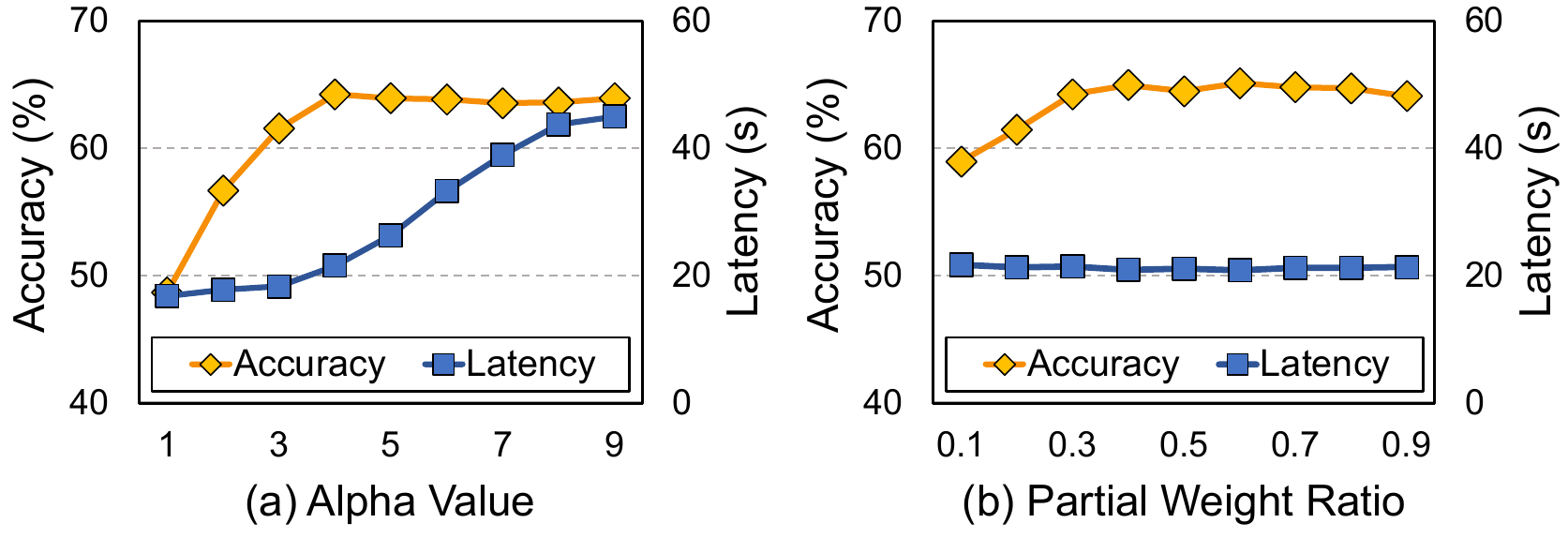}
  \caption{
    Accuracy and inference latency across (a) alpha values and (b) partial
    weight ratios.
  }
  \vspace{-0.15in}
  \label{fig:alpha_sweep}
\end{figure}

%%%%%%%%%%%%%%%%%%%%%%%%%%%%%%%%%%%%%%%%%%%%%%%%%%%%%%%%%%%%%%%%%%%%%%%%
\myparagraph{Threshold and Alpha.}
As discussed in~\ssecref{prefetching}, we load the KV cache of the tokens with
a speculated attention score greater than the threshold (i.e., the maximum
attention score minus alpha). Increasing alpha results in fetching more KV
entries to the GPU, thus increasing inference latency but also improving
accuracy.
\figref{alpha_sweep}(a) shows such trade-offs between accuracy and inference
latency for nine different alpha values with a partial weight ratio of 0.3. 
The results show that more KV entries are fetched and involved in attention
computation as alpha increases, thereby leading to better accuracy.
For the alpha values beyond 4, however, since most important tokens are already
included, the accuracy does not further increase, while the cost for KV
transfers and attention computation keeps increasing. This trend is similarly
observed in other models, and we thus opt for an alpha value of 4 or 5 to
strike a balance between inference latency and accuracy.

%%%%%%%%%%%%%%%%%%%%%%%%%%%%%%%%%%%%%%%%%%%%%%%%%%%%%%%%%%%%%%%%%%%%%%%%
\myparagraph{Partial Weight Ratio.}
\figref{alpha_sweep}(b) shows the accuracy and inference latency across
different partial weight ratios with an alpha value of 4. 
As shown in the figure, the amount of partial weights has a negligible impact
on inference latency because the cost for computing the speculated attention
score is relatively small. Note that the amount of KV cache to transfer is
\emph{not} related to the partial weight ratio.
However, increasing the partial weight ratio results in higher memory
consumption for partial weights and key cache (e.g., doubling the ratio doubles
the memory consumption overhead).
The accuracy also does not noticeably differ beyond a ratio of 0.3. In our
work, we opt for a partial weight ratio of 0.3 to achieve better accuracy while
considering memory consumption overhead.

%%%%%%%%%%%%%%%%%%%%%%%%%%%%%%%%%%%%%%%%%%%%%%%%%%%%%%%%%%%%%%%%%%%%%%%%
\putssec{}{Overhead}

\begin{figure}[t]
  \includegraphics[clip=True, trim=0in 0.1in 0in 0in, width=\linewidth]{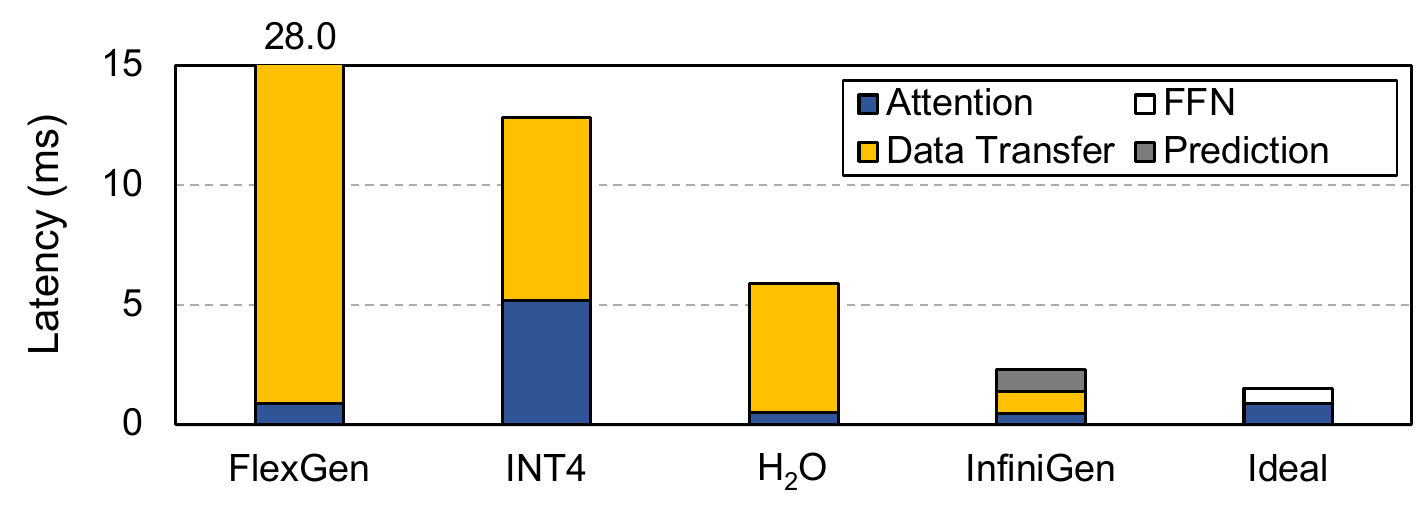}
  \caption{
    Latency breakdown of a Transformer block for OPT-13B with a sequence length
    of 2048 and a batch size of 8.
  }
  \vspace{-0.15in}
  \label{fig:block_latency_breakdown}
\end{figure}

%%%%%%%%%%%%%%%%%%%%%%%%%%%%%%%%%%%%%%%%%%%%%%%%%%%%%%%%%%%%%%%%%%%%%%%%
\noindent\textbf{Prefetching Overhead.}
\figref{block_latency_breakdown} shows the latency breakdown of executing a
single Transformer block for the OPT-13B model; FFN is not shown in the figure
for schemes other than Ideal because it is fully overlapped with data transfer
time. Ideal is a scenario where all the computations (i.e., attention and FFN)
are performed on the GPU without any data transfer between the CPU and GPU.
As shown in the results, the key performance bottleneck of FlexGen and \ho{} is
the data transfer overhead, which occupies 96.9\% and 91.8\% of the execution
time, respectively.
For INT4, due to the quantization and dequantization overhead, attention
computation also occupies a large portion of the execution time in addition to
the data transfer.
\name{}, on the other hand, considerably improves the inference speed over
FlexGen by reducing the amount of data transfer with our dynamic KV cache
prefetching. Furthermore, \name{} is only 1.52$\times$ slower than Ideal, while
others show 3.90$\times$-18.55$\times$ slowdowns.

%%%%%%%%%%%%%%%%%%%%%%%%%%%%%%%%%%%%%%%%%%%%%%%%%%%%%%%%%%%%%%%%%%%%%%%%
\myparagraph{Memory Consumption.}
\name{} uses the partial query weight and key cache for speculation.
For a ratio of 0.3, the sizes of the partial query weight and key cache are
only 2.5\% and 15\% of the total model parameters and total KV cache,
respectively. 
While we simply store them in the GPU during our experiments, we can manage the
storage overhead in various optimized ways if needed. For example, we can store
only the column indices of the partial query weight and retrieve the column
vectors from the full query weight matrix (which already resides in the GPU) as
needed for partial query projection. Additionally, we can place the partial key
cache in the CPU and perform speculation on the CPU after fetching the partial
query from the GPU.
Even a na\"ive method of lowering the partial ratio would likely still provide
better accuracy compared to other methods while reducing storage overhead.
In summary, by minimally sacrificing inference performance, we can greatly
reduce the storage overhead on the GPU if necessary.

\begin{figure}[t]
  \includegraphics[width=\linewidth]{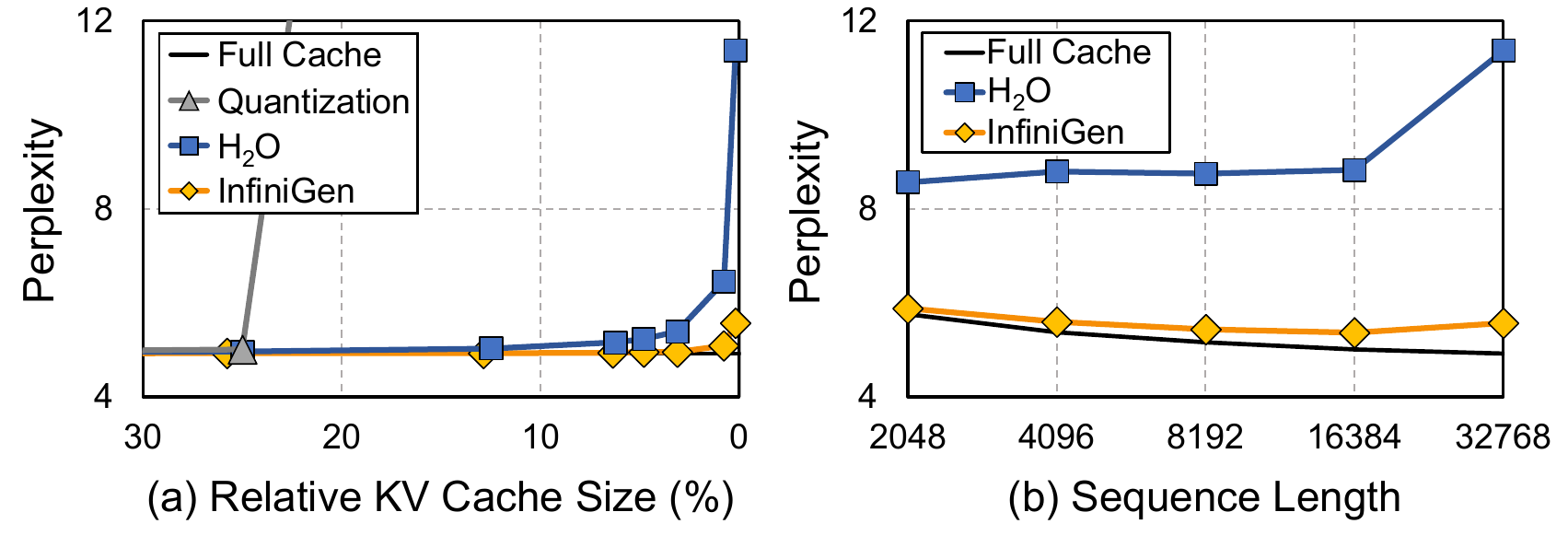}
  \caption{
    Perplexity of Llama-2-7B-32K across (a) relative KV cache sizes with a
    sequence length of 32768 and (b) sequence lengths while retaining 64
    tokens.  Lower is better.  Llama-2-7B-32K is a fine-tuned version capable
    of processing up to 32K tokens using position
    interpolation~\cite{che:won23}.  Quantization is omitted in (b) since the
    KV cache cannot be compressed below 6.25\% (i.e., 1 bit).
  }
  \vspace{-0.15in}
  \label{fig:scalability}
\end{figure}

%%%%%%%%%%%%%%%%%%%%%%%%%%%%%%%%%%%%%%%%%%%%%%%%%%%%%%%%%%%%%%%%%%%%%%%%%%%%%%%%%%
\putssec{}{Long Context Window}
\figref{scalability} shows the perplexity of the Llama-2-7B-32K model, which
can process up to 32K tokens, across the relative cache sizes and sequence
lengths.
We use the WikiText-2 dataset for the experiment.
As the context window size increases for future LLMs, the relative portion of
the KV cache that the GPU can retain would decrease due to the limited capacity
of GPU memory.

\figref{scalability}(a) shows that \name{} maintains perplexity levels close to
the full-cache baseline even as the relative KV cache size decreases, without
leading to a noticeable increase in perplexity even with much smaller cache
sizes.
In contrast, other methods increase perplexity compared to the full-cache
baseline and significantly diverge at certain sizes due to insufficient bit
widths for preserving adequate information on all keys and values
(Quantization) or the permanent removal of KV cache entries (\ho{}).
As shown in~\figref{scalability}(b), the perplexity gap between \name{} and
\ho{} widens for longer sequence lengths, which is likely to increase further
for sequence lengths beyond 32K. This implies that \name{} can scale to longer
sequences and better preserve model accuracy compared to others.

\begin{figure}[t]
  \includegraphics[width=\linewidth]{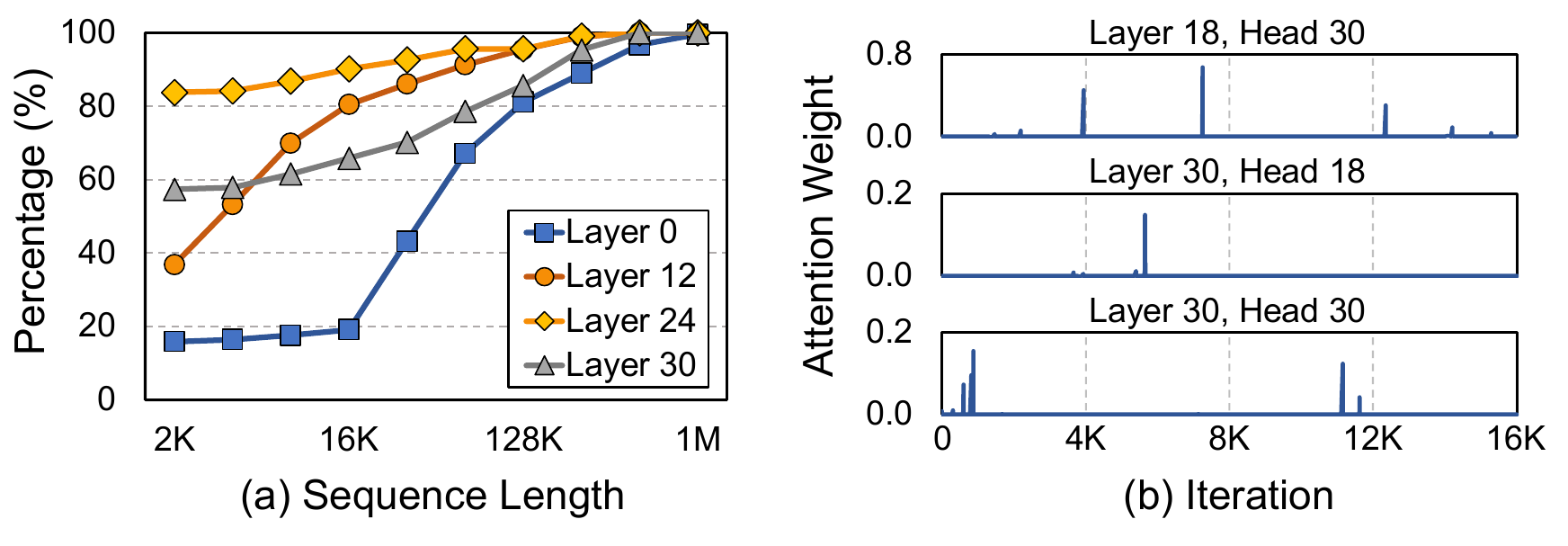}
  \caption{
    Analysis of 1 million tokens using Llama-3-8B-1048K. (a) Percentage of
    query tokens that attend to less than 1\% of the key tokens across the
    sequence lengths. (b) Attention weight of sampled key tokens from different
    layers and heads across the last 16K iterations out of 1 million tokens.
  }
  \vspace{-0.15in}
  \label{fig:1m}
\end{figure}

We further speculate on how \name{} would benefit in an era of million-token
context windows by analyzing a model capable of handling 1 million tokens. 
\figref{1m}(a) shows that the percentage of query tokens that attend to less
than 1\% of key tokens increases as the sequence length becomes longer. 
\name{} can adapt to this changing trend by dynamically adjusting the amount of
the KV cache to load, whereas prior fixed-budget/pruning approaches would not
easily adjust the effective KV cache size.
\figref{1m}(b) further shows that the attention weights of key tokens can
change across iterations; the sampled key tokens show sudden \emph{spikes}
after thousands of iterations with significantly low attention weights (e.g.,
the 7425{$^\textrm{th}$} iteration out of the last 16K iterations in Layer 18,
Head 30).
We observe that the prior approaches that permanently eliminate tokens while
they are unimportant could lose the critical contexts if they become important
again at later iterations. 
In contrast, \name{} can preserve model performance by keeping the temporarily
unimportant KV entries for potential future use.

\putsec{related}{Related Work}
\noindent \textbf{DNN Serving Systems.}
A systematic approach to enabling an efficient and fast model serving system is
an important topic that has been widely studied by both academia and industry.
Some prior works focus on distributed systems with predictable latency for
service-level objectives (SLOs)~\cite{cra:xin17,cra:gur20,guj:rez20,she:leq19}.
Other works improve parallelism and throughput of the system through
preemption~\cite{han:han22,zha:yup23}, fine-grained batching~\cite{cui:zha22,
yu:jeo22, fan:yan21}, or memory optimizations~\cite{dao:dan22, kwo:zhu23,
shi:yan23}. 

Several other works aim at achieving high throughput execution with limited GPU
memory by offloading parameters to secondary storage (e.g., CPU memory and
disk). Some of them build on CUDA Unified Memory~\cite{uvm} with
prefetching~\cite{jun:kim23, mar:bel18}, while others explicitly move tensors
in and out as needed for computation~\cite{hao:yir23, hua:gu20, yu:mar18,
hil:kha20, pen:shi20}. FlexGen~\cite{she:zhe23} is a recent LLM serving system
that enables high-throughput inference on a single GPU by offloading weights
and KV cache to CPU memory and disk.
\name{} is orthogonal to FlexGen and can work in conjunction with it to
efficiently offload and prefetch the KV cache.

%%%%%%%%%%%%%%%%%%%%%%%%%%%%%%%%%%%%%%%%%%%%%%%%%%%%%%%%%%%%%%%%%%%%%%%%
\myparagraph{KV Cache Management.}
vLLM~\cite{kwo:zhu23} mitigates the KV cache memory waste from fragmentation
and duplication. StreamingLLM~\cite{xia:tia23} enables LLMs to generate longer
sequence lengths than the trained ones. 
However, since neither vLLM nor StreamingLLM reduces the size of the KV cache,
data transfers still incur a significant overhead in offloading-based inference
systems.
\name{} complements the KV cache management to reduce the data transfer
overhead, which is a major bottleneck in offloading-based systems.

%%%%%%%%%%%%%%%%%%%%%%%%%%%%%%%%%%%%%%%%%%%%%%%%%%%%%%%%%%%%%%%%%%%%%%%%
\myparagraph{Efficient LLM Inference.}
There are lines of research that exploit \emph{quantization} or \emph{sparsity}
to make LLMs efficient through algorithmic methods~\cite{chi:gra19, kwo:kim22,
fra:ali23, xia:lin22, det:lew22} or hardware-software
co-design~\cite{ham:lee21, qu:liu22, lee:lee24, guo:tan23}.
Regarding sparsity, most algorithm-based works focus on reducing the model size
by exploiting the sparsity of weights.
Alternatively, \ho{} and Sparse Transformer~\cite{chi:gra19} leverage the
row-wise (i.e., token-level) sparsity in the KV cache by \emph{permanently}
removing certain KV entries.
On the other hand, most hardware-software co-design studies focus on relaxing
the quadratic computational complexity in the prefill stage by skipping
non-essential key tokens with the aid of specialized hardware. 
However, they often do not reduce memory access as they identify the important
key tokens only after scanning all the elements of the key tensors.

Kernel fusion~\cite{dao:dan22, kao:sub23} is another approach to mitigating the
quadratic memory overhead of attention in the prefill stage. \name{} can be
implemented with kernel fusion techniques to alleviate the overhead of KV cache
access in the decoding stage. 
To our knowledge, this is the \emph{first} work to enable efficient LLM
inference by prefetching only essential KV entries in offloading-based
inference systems.

\putsec{conclusion}{Conclusion}
The size of the KV cache poses a scalability issue in high-throughput
offloading-based inference systems, even surpassing the model parameter size.
Existing KV cache eviction policies show a large accuracy drop and do not
efficiently use the interconnect bandwidth when they are employed in
offloading-based LLM systems. We propose \name{}, an offloading-based dynamic
KV cache management framework that efficiently executes inference of large
language models. \name{} exploits the attention input of the previous layer to
speculatively prefetch the KV cache of important tokens. We manipulate the
query and key weights to make the speculation more efficient. \name{} shows
substantially shortened inference latency while preserving language model
performance. It also shows much better scalability regarding the batch size,
sequence length, and model size compared to prior solutions.

\section*{Acknowledgments}
We would like to thank the anonymous reviewers and our shepherd Petros Maniatis
for their valuable feedback.
This work was supported in part by a research grant from Samsung Advanced
Institute of Technology (SAIT) and by the artificial intelligence semiconductor
support program to nurture the best talents (No. RS-2023-00256081) supervised
by Institute for Information \& Communications Technology Planning \&
Evaluation (IITP).
The Institute of Engineering Research at Seoul National University provided
research facilities for this work.
Jaewoong Sim is the corresponding author.

\bibliographystyle{plain}
\bibliography{refs}

\end{document}